\documentclass{article}

\PassOptionsToPackage{numbers,sort&compress}{natbib}
\usepackage[main, final]{neurips_2026}

\usepackage[utf8]{inputenc} 
\usepackage[T1]{fontenc}    
\usepackage{hyperref}       
\usepackage{url}            
\usepackage{booktabs}       
\usepackage{amsfonts}       
\usepackage{nicefrac}       
\usepackage{microtype}      
\usepackage{xcolor}         
\usepackage{microtype}
\usepackage{graphicx}
\usepackage{subcaption}
\usepackage{booktabs} 
\usepackage{multirow}
\usepackage{makecell}
\usepackage{comment}
\usepackage{tabularx}
\usepackage{float}
\newcolumntype{Y}{>{\centering\arraybackslash}X}
\usepackage{amsmath}
\usepackage{amssymb}
\usepackage{mathtools}
\usepackage{amsthm}
\usepackage{array}
\usepackage{wrapfig}
\usepackage{enumitem}
\usepackage{wrapfig}

\usepackage{hyperref}

\usepackage[capitalize,noabbrev]{cleveref}
\usepackage{placeins}

\theoremstyle{plain}
\newtheorem{theorem}{Theorem}[section]

\theoremstyle{definition}
\newtheorem{definition}[theorem]{Definition}

\theoremstyle{remark}

\newcommand{\spara}[1]{\vspace{1ex}\noindent\textbf{#1}}
\newcommand{\algshort}{SIMBAD}

\title{SIMBAD: Spatio-Temporal Traffic Forecasting \\Robust to Aperiodicity}

\author{
  Daniel Y. Lee \\
  Korea University \\
  \texttt{theghks3@korea.ac.kr}
  \And
  Seungwon Shin \\
  Korea University \\
  \texttt{mengchi501@korea.ac.kr}
  \And
  Seunghoon Han \\
  Chungnam National University \\
  \texttt{tmdgns129@g.cnu.ac.kr}
  \And
  Sungsu Lim\textsuperscript{$\dagger$} \\
  Chungnam National University \\
  \texttt{sungsu@cnu.ac.kr}
  \And
  Susik Yoon\textsuperscript{$\dagger$} \\
  Korea University \\
  \texttt{susik@korea.ac.kr}
}

\begin{document}

\maketitle

\begingroup
\renewcommand{\thefootnote}{\fnsymbol{footnote}}
\footnotetext[2]{Corresponding authors.}
\endgroup

\begin{abstract}
  Accurate traffic prediction is essential for urban planning but remains challenging due to the irregular patterns caused by unexpected events. While Spatio-Temporal Graph Neural Networks (STGNNs) effectively model periodic data, they often struggle with these aperiodic fluctuations. To address this, we propose \algshort{}, which improves generalization on irregular patterns via two key mechanisms: 1) adaptive regulation of periodic signals to prioritize recent signals, and 2) dynamic spatial modeling that adjusts influence between connected nodes based on contextual relations. Built upon simple similarity metrics to adapt to aperiodic patterns, \algshort{} outperforms existing methods on real-world benchmark datasets, demonstrating superior performance in predicting unseen and irregular traffic events.
\end{abstract}

\section{Introduction}
\begin{figure}[H]
    \centering
    \begin{subfigure}{0.48\textwidth}
        \centering
        \includegraphics[width=\linewidth]{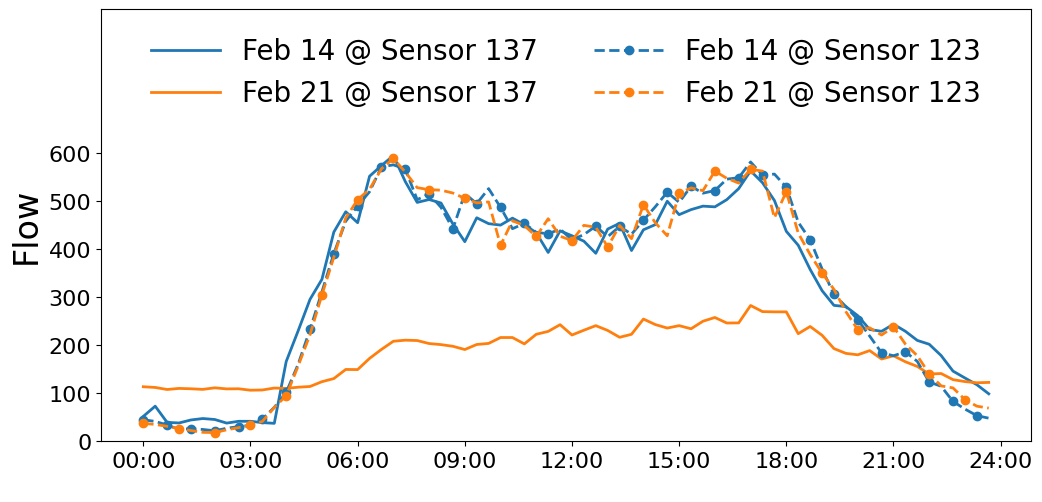}
        \caption{}
        \label{fig1a}
    \end{subfigure}
    \begin{subfigure}{0.25\textwidth}
        \centering
        \includegraphics[width=\linewidth]{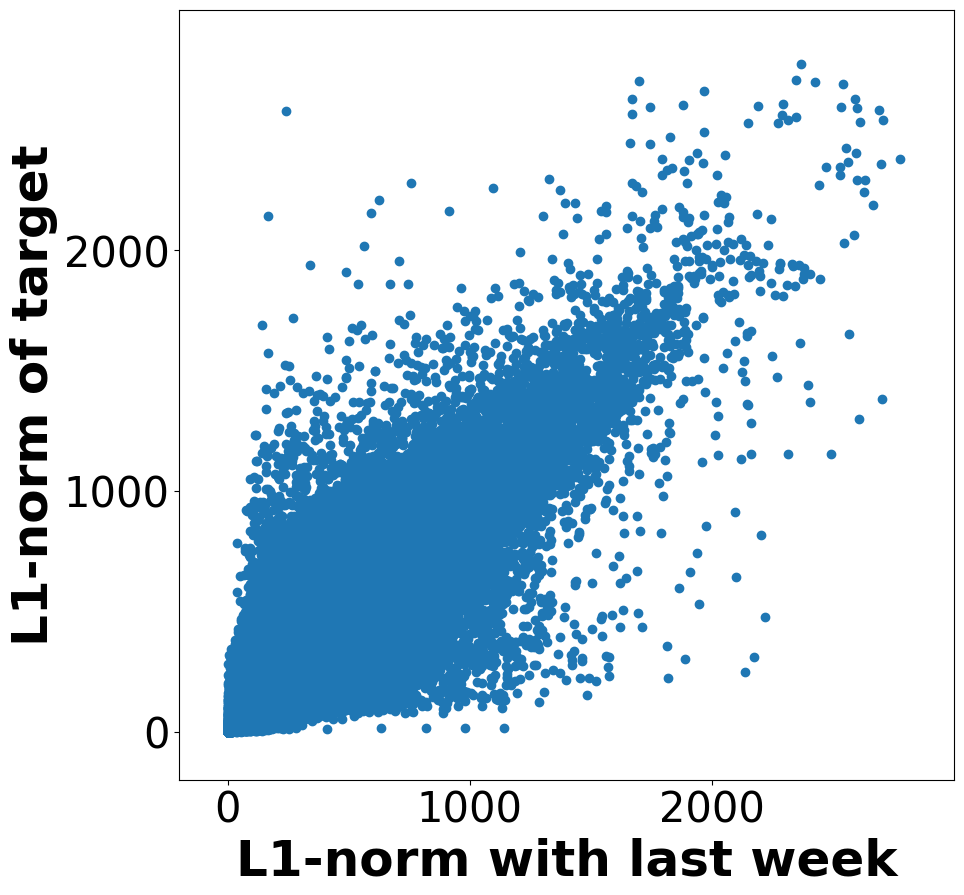}
        \caption{}
        \label{fig1b}
    \end{subfigure}
    \begin{subfigure}{0.24\textwidth}
        \centering
        \includegraphics[width=\linewidth]{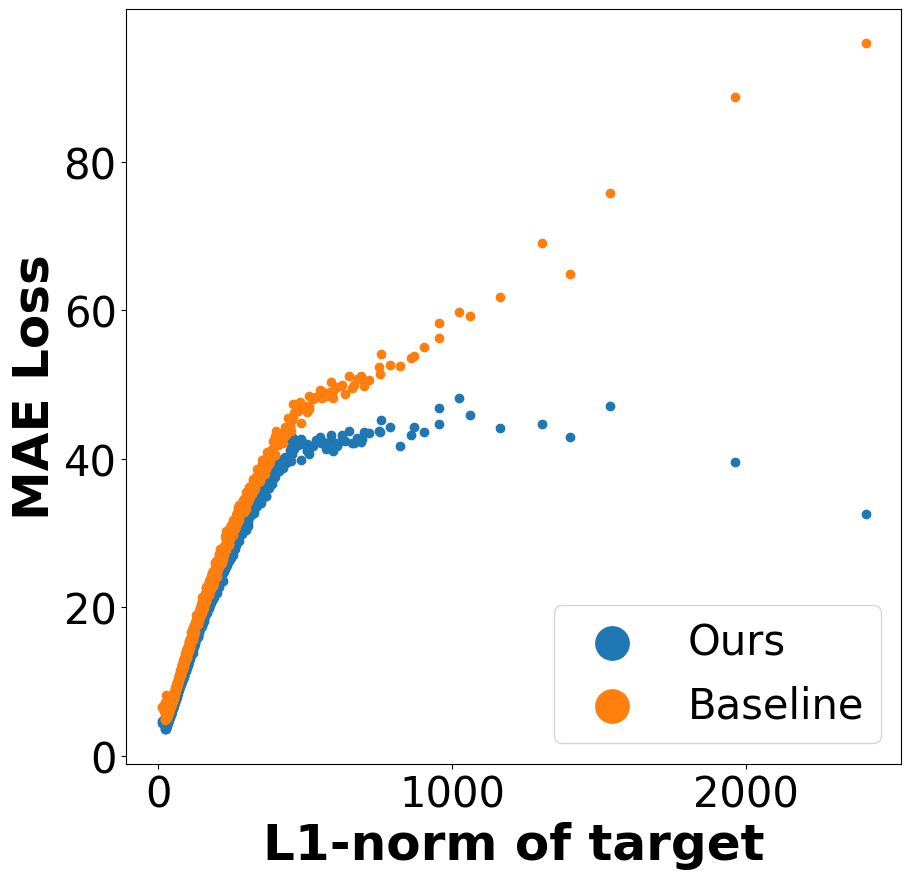}
        \caption{}
        \label{fig1c}
    \end{subfigure}
    \caption{Motivating examples on PEMS04 dataset. (a) Varying periodic traffic patterns among connected nodes. (b) Correlation between L1-norms of input and target sequences with respect to their last week counterparts. (c) Performance comparison under growing aperiodicity.}
    \vspace{-0.2cm}
    \label{fig_motive}
\end{figure}
Accurate spatio-temporal forecasting is a prominent research area due to its broad applicability, particularly in traffic flow forecasting~\cite{trafficforecasting} for Intelligent Transportation Systems (ITS)~\cite{ITS}. Despite its practical importance, accurate spatio-temporal traffic forecasting remains challenging due to inherent complex dependencies and irregularities. As shown in Fig. \ref{fig_motive}(a), traffic data from interconnected roads often exhibits heterogeneous patterns, such as temporal variations between periodic intervals (e.g., sensor 137) and spatial variations between neighboring nodes (e.g., sensors 123 and 137).

To address this aperiodicity and achieve accurate forecasting, recent studies~\cite{stfgnn, pdformer} have utilized semantic adjacency matrices based on Dynamic Time Warping (DTW)~\cite{dtw}. However, the high computational complexity of DTW limits its scalability, motivating the use of more efficient alternatives like Minkowski distance \cite{minkowski}. The use of historical information is grounded in extensive empirical evidence on the recurrent nature of human mobility and urban traffic. Prior studies have shown that human mobility exhibits highly reproducible spatial and temporal patterns~\cite{mobility, limit_mobility}, while urban congestion exhibit substantial day-to-day regularity~\cite{daytoday}. Notably, as shown in Fig. \ref{fig_motive}(b), we observe that the simple L1 distance between current and historical data (e.g., last week) is still sufficient to show a strong correlation between the corresponding distance of the input sequence and that of the target sequence. It suggests that lightweight distance metrics can effectively quantify underlying regularities in traffic patterns.

Besides, another line of research leverages the strong periodicity of traffic data, such as recurring peak hours on weekdays. For instance, ST-ResNet~\cite{stresnet} explicitly models temporal closeness, period and trend, while ASTGCN~\cite{astgcn} incorporates weekly, daily, and recent sequences to better predict periodic patterns. Subsequent works~\cite{traffictransformer, stae} further advanced this by introducing periodic positional encodings, enabling models to represent recurring temporal regularities at each position more effectively. Despite these successes, reliance on historical sequences can degrade performance. Revisiting Fig. \ref{fig_motive}(a), traffic patterns often vary even on the same day across different weeks. This discrepancy at a certain timestep tends to propagate to its future states as empirically quantified in Fig. \ref{fig_motive}(b), making such aperiodic historical sequences less informative for predicting the target sequences. Consequently, existing models that overlook these variations experience significant performance degradation as the aperiodicity of the target sequence increases.

To address these challenges, we propose \textbf{\algshort{}}, a \textit{simple yet effective} \underline{SIM}ilarity-\underline{B}ased \underline{A}periodicity-robust forecasting metho\underline{D}. We explicitly exploit similarity between historical and input traffic patterns, enabling adaptive learning of spatio-temporal dependencies. Specifically, we demonstrate that the lightweight L1 distance between current and past periodic data can sufficiently guide a graph neural network, enabling \algshort{} to adaptively consult past traffic information and capture spatio-temporal patterns. In addition, we adopt the uncertainty measure~\cite{uncertainty} to regulate noise in the L1 distance when quantifying the discrepancy observed between current and past traffic patterns. \algshort{} then regulates information flow by modulating spatial interactions between nodes and selectively incorporating historical data that aligns with recent dynamics, ensuring robust forecasting under aperiodic conditions. Consequently, \algshort{} is particularly more effective in forecasting aperiodic patterns under unexpected sporadic events. As shown in Fig. \ref{fig_motive}(c), \algshort{} exhibits significantly lower performance degradation compared to the seminal periodicity-aware method~\cite{astgcn} as the aperiodicity of the target sequence increases.

In brief, our contributions are summarized as follows:
\begin{itemize}[leftmargin=15pt, noitemsep]
\item We propose a simple yet effective similarity-based spatio-temporal forecasting approach, overcoming the practical challenges of handling sporadic, irregular patterns.
\item We present a graph neural network framework \algshort{}, employing an Aperiodicity-Aware Modulation Layer and a Spatio-Temporal Modeling layer, to dynamically adjust the flow of temporal information between connected nodes. Furthermore, incorporating the proposed Aperiodicity-Aware Modulation Layer into existing baselines improves their predictive performance, demonstrating its effectiveness as a transferable module. The source code is publicly available at \url{https://github.com/theghks3/SIMBAD}.
\item Extensive experiments on four widely used benchmark datasets, and three supplementary datasets with different regions or large-scale, demonstrate that our model not only achieves strong overall performance but also excels at predicting unseen, irregular traffic patterns, achieving $4.12\%$ of accuracy improvement on average over the best baseline.
\end{itemize}
\section{Preliminary}
We formalize the prediction problem in a traffic network and the notations used throughout the paper.

\begin{definition}
    The \textit{traffic network} is represented as a graph $\mathcal{G} = (\mathcal{V},\mathcal{E},\mathbf{A})$, where $\mathcal{V}$ is a set of $N$ nodes, $\mathcal{E} \subseteq \mathcal{V} \times \mathcal{V}$ is a set of edges, and $\mathbf{A} \in \mathbb{R}^{N \times N}$ is the weighted adjacency matrix of graph $\mathcal{G}$.
\end{definition}

Given previous traffic observation of $T$ timesteps, the typical goal of traffic forecasting is to predict future traffic flow of $T^{\prime}$ timesteps with input $X \in \mathbb{R}^{N \times f \times T}$ by learning a mapping function $F$:
\begin{equation}
[X_{t-T+1}, \cdots, X_{t};\mathcal{G}] \underset{F}\longrightarrow [X_{t+1}, \cdots, X_{t+T^{\prime}}].
\end{equation}

\begin{definition}
    Given a timestep $t$, we define two auxiliary \textit{traffic windows} of length $T$, which represent the preceding and subsequent context window of $t$, respectively:
    \begin{equation}
    \begin{aligned}
        \mathcal{X}_{t}^{T-} &= [X_{t-T+1}, \cdots, X_{t}] \in \mathbb{R}^{N \times f \times T} \text{ and }
        \mathcal{X}_{t}^{T+} = [X_{t+1}, \cdots, X_{t+T}] \in \mathbb{R}^{N \times f \times T}.
    \end{aligned}
    \label{window}
    \end{equation}
    
    Then, we introduce the two types of inputs at a given timestep $t$:
\begin{itemize}[leftmargin=15pt, noitemsep]
\item \emph{Recent input} represents the previous observation of $T$ timesteps preceding $t$:
    \begin{equation}
        \mathcal{X}_{R} = \mathcal{X}_{t}^{T-}=[X_{t-T+1}, \cdots, X_{t}] \in \mathbb{R}^{N \times f \times T}.
    \end{equation}

\item \emph{Periodic input} represents historical windows of length $T$ starting at various periodic intervals prior to the onset of the target sequence:
    \begin{equation}
    \begin{split}
    \mathcal{X}_H^{(k)} = \mathcal{X}_{t-h_{k}}^{T+} &= \left[ X_{t - h_k + 1}, \dots, X_{t - h_k + T} \right] \in \mathbb{R}^{N \times f \times T},
    \end{split}
    \label{periodic_input}
    \end{equation}
    where $\mathcal{H} = \{h_1, h_2, \dots, h_K\}$ is a set of periodic intervals. For example, with $d$ days and the number $t'$ of timesteps in a day, $h_k = d \times t'$ and $\mathcal{X}_H^{(k)}$ corresponds to the historical window observed $d$ days prior to the target sequence.
\end{itemize}

    Given these inputs, we aim to learn the mapping function $F$ which maps the multiscaled historical information to future traffic states:
    \begin{equation}
        [\mathcal{X}_R, \mathcal{X}_H ; \mathcal{G}] \underset{F}\longrightarrow [X_{t+1}, \cdots, X_{t+T^{\prime}}],\: \mathcal{X}_{H} = \text{Concat}(\mathcal{X}_{H}^{(1)}, \mathcal{X}_{H}^{(2)}, \cdots, \mathcal{X}_{H}^{(K)}).
    \end{equation}
\end{definition}

\section{SIMBAD Framework}

\begin{figure*}[t]
\centering

\begin{subfigure}{\textwidth}
\centering
\includegraphics[width=\linewidth]{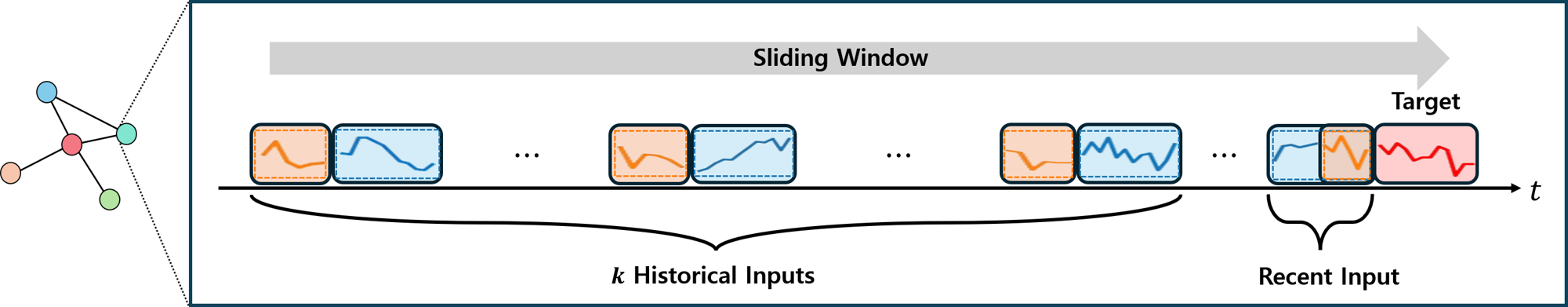}
\caption{Input segments of SIMBAD.}
\end{subfigure}


\begin{subfigure}{\textwidth}
\centering
\includegraphics[width=\linewidth]{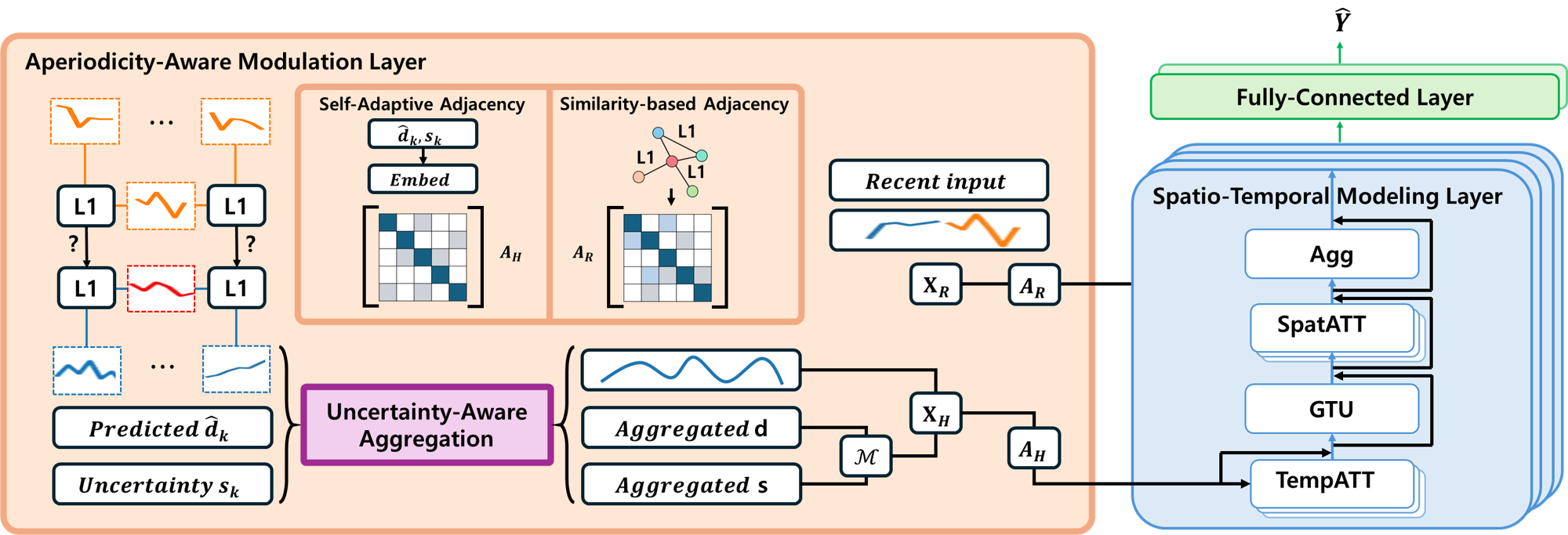}
\caption{Main components of SIMBAD.}
\end{subfigure}
\vspace{-0.3cm}
\caption{Overall framework of \algshort{}.}
\vspace{-0.2cm}
\label{fig_overall}
\end{figure*}

Figure \ref{fig_overall} shows the overall framework of \algshort{} with two key layers. The aperiodicity-aware modulation layer exploits temporal and spatial similarities to adaptively modulate historical inputs with uncertainty and semantic spatial dependencies; the spatio-temporal modeling layer models the regulated temporal and spatial signals to reflect spatio-temporal dynamics in target prediction.

\subsection{Aperiodicity-Aware Modulation Layer}

\subsubsection{Similarity-driven Temporal Regulation with Uncertainty}
\algshort{} regulates the periodic inputs based on the similarity with the recent input by \textit{L1-norm}:
\begin{equation}
    \delta(\mathcal{X},\mathcal{Y}) = \sum_{i=1}^{T} \left|X_{i}-Y_{i}\right| \in \mathbb{R}^{N \times f \times 1},
\label{l1}
\end{equation}
where $\mathcal{X}$ and $\mathcal{Y}$ represent two different traffic windows.

Specifically, we quantify the relevance of periodic input $\mathcal{X}_{H}^{(k)}$ to the current state based on the similarity of the current traffic sequence and the target sequence to each corresponding past periodic sequences, respectively:
\begin{equation}
    d_{k}^{-} = \delta(\mathcal{X}_{t}^{T-}, \mathcal{X}_{t-h_{k}}^{T-}), d_{k}^{+} = \delta(\mathcal{X}_{t}^{T+}, \mathcal{X}_{t-h_{k}}^{T+}).
    \label{input_target_sim}
\end{equation}

To address the discrepancy between $d_{k}^{-}$ and $d_{k}^{+}$, we model the \emph{aleatoric uncertainty} caused by noise inherent in the similarity measurement inspired by the prior work~\cite{uncertainty}. We first encode $d_{k}^{-}$ into a latent space and train a lightweight predictor and uncertainty head to estimate $\hat{d}_{k} \in \mathbb{R}^{N \times f \times 1}$ and the log variance $s_{k} = \text{log} \hat{\sigma}_{k}^{2} \in \mathbb{R}^{N \times f \times 1}$, respectively. The formulated uncertainty enables the model to be more robust to noisy data by mitigating the effect of erroneous signals. Beyond robust similarity estimation, the predicted similarity and uncertainty are explicitly incorporated into the model to modulate how periodic input is utilized.

\spara{Uncertainty-Guided Aggregation.} We estimate a similarity-based probability distribution to adaptively aggregate multiple periodic inputs based on the predicted similarity $\hat{d}_{k}$. Specifically, we incorporate the uncertainty as a confidence signal to down-weight the uncertain similarity signal:
\begin{equation}
    p_{k} = p(k|\{\tilde{d}_{j}\}_{j=1}^{k}) = \frac{\exp(-\tilde{d}_{k} / \tau)}{\sum_{j} \exp(-\tilde{d}_{j} / \tau)},  \: \tilde{d}_{k} = \hat{d}_{k} \times \hat{s}_{k}, \: \hat{s}_{k} = \phi(-s_{k}), 
    \label{distribution}
\end{equation}
where $\phi$ is a sigmoid function and $\tau$ is a temperature. Then, an aggregated periodic input is estimated:
\begin{equation}
    \tilde{\mathcal{X}}_{H} = \mathbb{E}_{k \sim p_{k}} [\mathcal{X}_{H}^{(k)}] \in \mathbb{R}^{N \times f \times T}.
    \label{aggregated}
\end{equation}


\algshort{} further constructs a masking matrix to regulate the inclusion of the historical information. Since the periodic input $\mathcal{X}_{H}$ is aggregated under the probability distribution in Eq. \ref{distribution}, we expect the input to follow the estimated similarity and confidence as:
\begin{equation}
    \mathbf{d} = \mathbb{E}_{k \sim p_{k}}[\hat{d}_{k}] \in \mathbb{R}^{N \times f \times 1}, \quad \mathbf{s} = \mathbb{E}_{k \sim p_{k}}[\hat{s}_{k}] \in \mathbb{R}^{N \times f \times 1}.
\end{equation}
Then, we construct a mask with a \emph{learnable} threshold $\lambda$:
\begin{equation}
    \mathcal{M} = \phi (\lambda - \mathbf{d}) \times \mathbf{s} \in \mathbb{R}^{N \times f \times 1}.
    \label{mask}
\end{equation}

We then multiply the constructed masking matrix by the aggregated input to regulate the inclusion of $\mathcal{X}_{H}$ of each node and obtain the final historical input:
\begin{equation}
    \mathbf{X}_{H} = \tilde{\mathcal{X}}_{H} \odot \mathcal{M} \in \mathbb{R}^{N \times f \times T},
    \label{final_his}
\end{equation}
where $\odot$ is the Hadamard product. 

In this way, \algshort{} dynamically controls periodic input selection and feature masking, while the uncertainty makes it remain robust under irregular and noisy traffic conditions.

\subsubsection{Similarity-driven Semantic Spatial Regulation}
In a traffic network, the connectivity information between sensors plays a critical role in modeling information propagation from a spatial perspective. Prior works~\cite{stgcn, astgcn,dcrnn} model the spatial information using Chebyshev spectral filter~\cite{chebyshev} or incorporate Gaussian kernel-based adjacency~\cite{gaussian}, which assumes that spatial correlations decay with distance. However, in real-world traffic scenarios, fixed states such as simple connections or distances do not necessarily reflect the true spatial dependencies among sensors, as they can vary depending on traffic conditions and temporal context. To address this limitation, we construct similarity-driven semantic adjacency matrices that can dynamically regulate spatial information.

\spara{Self-adaptive Adjacency Matrix.} An actual connectivity between nodes may not necessarily lead to their spatial dependency. Since the historical input in Eq.~\ref{final_his} represents the most contextually relevant periodic state, it is important to group nodes that exhibit similar periodic behaviors. Thus, we define semantically meaningful dependencies, rather than relying on fixed connections.

Past works~\cite{dcrnn,gwnet} proposed a self-adaptive adjacency matrix to guide the model to discover hidden spatial dependencies by randomly initializing node embedding dictionaries. Instead of random initialization, we initialize the node embeddings based on the reliability of historical information using $\hat{d}_{k}$ and $s_{k}$. Specifically, we initialize each node according to its similarity and confidence to the current state as $E \in \mathbb{R}^{N \times 2K}$. Then, with learnable parameters $W_{1}, W_{2} \in \mathbb{R}^{2K \times C}$, two node embeddings are initialized as:
\begin{equation}
    E = \phi(\text{Concat}(-\hat{d}_{k} || -s_{k})), \quad E_{1} = EW_{1}, \quad E_{2} = EW_{2}.
\end{equation}
Finally, the self-adaptive adjacency matrix is constructed as:
\begin{equation}
    \mathbf{A}_{H} = \text{Softmax}(\text{ReLU}(E_{1}E_{2}^\intercal)) \in \mathbb{R}^{N \times N},
\end{equation}
where the ReLU activation function eliminates weak connections, and the Softmax operation normalizes the self-adaptive adjacency matrix. In this way, the model is able to learn semantically meaningful connections based on the reflection of historical information to current traffic information.

\spara{Similarity-based Adjacency Matrix.} Since the prediction target corresponds to the immediate future following the current state, the actual connectivity reflected in the recent input plays a crucial role in determining how information propagates across nodes, encouraging the need to use real connectivity relations. However, relying solely on connectivity information does not model the potential variations in information among nodes. We further weight the pairwise node connections dynamically depending on their recent input similarities based on the L1 distance:
\begin{equation}
    S_{ij} = ||(\mathcal{X}_{t}^{T-})_{i} - (\mathcal{X}_{t}^{T-})_{j}||_{1},
\end{equation}
where $S_{ij}$ denotes the L1-distance between nodes $i$ and $j$. 

Since this spatial similarity is derived from the same sequence representation of temporal similarity in Eq.~\ref{input_target_sim}, we weight the spatial connections using the same threshold in Eq.~\ref{mask} to maintain consistent interpretations of similarity across temporal and spatial modeling. Thus, given the predefined adjacency matrix $A_{R}$, we weight each connection between node $i$ and $j$ as:
\begin{equation}
    (A_{R})_{ij} = \phi(\lambda - S_{ij}) \in \mathbb{R}^{N \times N}.
\end{equation}
Finally, we normalize the adjacency matrix for learning stability:
\begin{equation}
    \mathbf{A}_{R} = D^{-\frac{1}{2}}A_{R}D^{-\frac{1}{2}}, \quad D_{ii} = \sum_{j}{(A_{R})_{ij}}.
\end{equation}

\subsection{Spatio-Temporal Modeling Layer}
A spatio-temporal modeling layer contains three key components: 1) Self-Attention Block where vanilla transformers capture both spatial and temporal correlations, 2) Spatial Convolution Block which handles spatial dependencies from $k$-hop neighbors with various aggregation methods, and 3) Gated Temporal Convolution Block which focuses on catching temporal dependencies from nearby timesteps in a node. To effectively process the hidden periodic information, first $l_{h}$ layers process the historical data, followed by $l_{r}$ layers which process the recent data. Let hidden dimension be $f^{\prime}$.

\spara{Self-Attention Block.} We apply vanilla transformers to capture traffic relations in various perspectives. Given the input $\mathbf{X} \in \mathbb{R}^{N \times f^{\prime} \times T}$, we apply self-attention to capture dependencies within data. We first compute the query $(Q)$, key $(K)$ and value $(V)$ matrices as:
\begin{equation}
    Q = \mathbf{X}H_{Q}, K=\mathbf{X}H_{K}, V=\mathbf{X}H_{V},
\end{equation}
where $H_{Q}, H_{K}, H_{V} \in \mathbb{R}^{f^{\prime} \times f^{\prime}}$ are learnable parameters. Then, we compute the attention score matrix $A$ and apply the attention score to the value matrix as $\mathbf{Z} = AV$. We apply this calculation along spatial and temporal axes separately. Thus, each attention scores are $A^{sp}\in \mathbb{R}^{T \times N \times N}$ and  $A^{te} \in \mathbb{R}^{N \times T \times T}$ which contains spatial relations of different timesteps and temporal relations for different nodes, respectively. The output of each attention block is defined as:
\begin{equation}
    \text{SpatATT}(\mathbf{X}), \text{TempATT}(\mathbf{X}) \in \mathbb{R}^{N \times f^{\prime} \times T}.
\end{equation}

\spara{Spatial Convolution Block.} Graph convolution is an essential operation in extracting a node's features within the structural information. Studies~\cite{spectral,chebyshev} proposed a graph convolutional method based on spectral graph theory and \cite{cheby_first} proposed a first approximation of Chebyshev spectral filtering. These spectral-based approaches enable the modeling of spatial dependencies by performing convolution in the graph Fourier domain. A single-hop graph convolution is defined with the normalized adjacency matrix $\tilde{\mathbf{A}} \in \mathbb{R}^{N \times N}$, input $\mathbf{X} \in \mathbb{R}^{N \times f^{\prime}}$, and learnable parameter $\mathbf{W} \in \mathbb{R}^{f^{\prime} \times f^{\prime}}$:
\begin{equation}
    \mathbf{X}_{\text{agg}} = \tilde{\mathbf{A}} \mathbf{X} \mathbf{W} \in \mathbb{R}^{N \times f^{\prime}}.
\end{equation}
Then, the spatial convolution block is defined to aggregate information from $K$-hop neighbors: 
\begin{equation}
    \text{AGG}(\mathbf{X}, \tilde{\mathbf{A}}) = \sum_{k=0}^{K}{\tilde{\mathbf{A}}^{k}\mathbf{X}\mathbf{W}_{k}}.
\end{equation}

\spara{Gated Temporal Convolution Block.} In addition to capturing local dependencies, it is crucial to model temporal dynamics by merging information across neighboring timesteps. To address this, we apply 1D convolution along the temporal axis to learn local temporal patterns. However, standard 1D convolution may treat all timesteps equally and fail to distinguish between irregular timesteps. To overcome this limitation we incorporate a Gated Tanh unit (GTU)~\cite{gtu} to capture the temporal dynamic information of traffic flow data. GTU in time dimension can be defined as:
\begin{equation}
    \text{GTU}(\mathbf{X}) = \phi(\Theta_{1} * \mathbf{X} + a) \odot \text{tanh}(\Theta_{2} * \mathbf{X} + b),
\end{equation}
where $\Theta_{1}$ and $\Theta_{2}$ are two independent 1D convolution operations.

\subsection{Optimization}
Given input $\mathbf{X}$ and adjacency matrix $\mathbf{A}$, the output of \algshort{} layer $g(\cdot)$ is obtained by stacking the temporal modeling, then spatial modeling, respectively:
\begin{equation}
    g(\mathbf{X}, \mathbf{A}) = \text{AGG}(\text{SpatATT}(\text{GTU}(\text{TempATT}(\mathbf{X}))), \mathbf{A}).
\end{equation}
As mentioned, first $l_{h}$ layers process the historical data to obtain $g^{l_{h}}(\mathbf{X}_{H}, \mathbf{A}_{H})$. Since the dissimilar nodes of historical input are masked as in Eq.~\ref{final_his}, the recent input of corresponding nodes must be emphasized for accurate prediction. Thus, we add the whole recent input and the recent input of corresponding nodes through $\mathcal{M}$:
\begin{equation}
\begin{split}
\mathbf{X}^{\prime}_{R}
&= \text{LayerNorm}\Bigl(
g^{l_{h}}(\mathbf{X}_{H}, \mathbf{A}_{H}) + \mathbf{X}_{R} + (1 - \mathcal{M}) \odot \mathbf{X}_{R}
\Bigr).
\end{split}
\end{equation}

Then, we generate the final output
\begin{equation}
    \hat{Y} = \text{FC}(\text{FC}(g^{l_{h} + l_{r}}(\mathbf{X}^{\prime}_{R}, \mathbf{A}_{R}))),
\end{equation}

where $\hat{Y} \in \mathbb{R}^{N \times f_{o} \times T^{\prime}}$ is the horizon of prediction. In our case, the output dimension $f_{o}$ is $1$.

We supervise the uncertainty $s_{k}$ derived from discrepancy between equations in Eq.~\ref{input_target_sim}~\cite{uncertainty}: 
\begin{equation}
    \mathcal{L}_{u} = {\frac{1}{K}} \sum_{k} {{1} \over {2}} \text{exp}(-s_{k}) ||d_{k}^{+} - \hat{d}_{k}||^{2} + {{1} \over {2}} s_{k}.
    \label{unc_loss}
\end{equation}

To predict the future timesteps $Y \in \mathbb{R}^{N \times f \times T^{\prime}}$, we minimize the $L1$ loss:
\begin{equation}
    \mathcal{L}_{traffic} = \sum_{T=t+1}^{t+T^{\prime}} \sum_{n=1}^{N}{\left|y_{T}^{n} - \hat{y}_{T}^{n} \right|}.
\end{equation}

Finally, \algshort{} aims to jointly minimize the combined loss with a regularization factor $\lambda_{u}$:
\begin{equation}
    \mathcal{L} = \mathcal{L}_{traffic} + \lambda_{u} \cdot \mathcal{L}_{u}.
\end{equation}
\section{Experiments}
\subsection{Experiment Settings}
\spara{Datasets and Evaluation Metrics}
We verify the performance of \algshort{} mainly on four widely used benchmark datasets~\cite{stsgcn}, PEMS03, PEMS04, PEMS07, and PEMS08. Traffic flow is measured every 5 minutes, resulting in 288 timesteps in a day, and standard normalization is applied to the input data. Details of datasets are given in Table \ref{tab:dataset}. 
\begin{wraptable}{r}{0.50\linewidth}
\vspace{-2pt}
\small
\centering
\caption{Summary of datasets.}
\vspace{-4pt}
\label{tab:dataset}
\begin{tabular}{lccc}
\toprule
\textsc{Dataset} & \textsc{Nodes} & \textsc{Edges} & \textsc{Timesteps}\\
\midrule
PEMS03 & 358 & 547 & 26,208 \\
PEMS04 & 307 & 340 & 16,992 \\
PEMS07 & 883 & 866 & 28,224 \\
PEMS08 & 170 & 295 & 17,856 \\
\bottomrule
\end{tabular}
\vspace{-10pt}
\end{wraptable}
We also employed three supplementary datasets (METR-LA, PEMS-BAY, and LargeST) with the evaluation results provided in Appendix \ref{ap:additional_datasets}.
We used the three most frequently used evaluation metrics in traffic forecasting: root mean squared error (RMSE), mean absolute error (MAE), and mean absolute percentage error (MAPE). We calculate the error based on the average performance of all predicted horizons on the test sets.

\spara{Baselines}
We compared \algshort{} with 11 baselines categorized into three groups: (1) GNN-based (\textit{GWNet}~\cite{gwnet}, \textit{DCRNN}~\cite{dcrnn}, \textit{AGCRN}~\cite{agcrn}, \textit{STSGCN}~\cite{stsgcn}, \textit{DSTAGNN}~\cite{dstagnn}, and \textit{ASTGCN}~\cite{astgcn}), (2) Transformer-based (\textit{STAEformer}~\cite{stae}, \textit{STWave}~\cite{stwave}, \textit{DTRformer}~\cite{DTRformer}, and \textit{LHMformer}~\cite{lhmformer}), and (3) Memory-based (\textit{STIDGCN}~\cite{stidgcn}). More details are given in Appendix \ref{ap:baseline}.

\spara{Implementation Details}
To align with recent studies, we split datasets into train, validation, and test sets in a $6:2:2$ ratio. We train our model using Adam optimizer with a learning rate of 0.001 and the batch size of 64. We use 2 weeks, 1 day, and 1 hour of historical data for the prediction of the next 12 timesteps. The hidden dimension $f^{\prime}$ is 64, and the number of layers $l_{h}$ and $l_{r}$ are both set to 2. Temperature $\tau$ is set as 0.2 and regularization factor $\lambda_{u}$ is 0.1. We use 4 heads for self-attention transformers. The training epoch is set to 100 with an early-stop approach for which the validation error does not improve within 30 epochs. All baselines used the default hyperparameters suggested in the original work or were fine-tuned to achieve the best scores.

\subsection{Overall Performance Results}

\begin{table*}[h]
\centering
\caption{Performance comparison and ablation study results on PEMS datasets.}
\label{tab:overall_ablation}
\resizebox{\textwidth}{!}{
\begin{tabular}{@{}l ccc ccc ccc ccc@{}}
\toprule
\textsc{Model}
& \multicolumn{3}{c}{PEMS03}
& \multicolumn{3}{c}{PEMS04}
& \multicolumn{3}{c}{PEMS07}
& \multicolumn{3}{c}{PEMS08} \\
\cmidrule(lr){2-4}
\cmidrule(lr){5-7}
\cmidrule(lr){8-10}
\cmidrule(lr){11-13}
\textsc{Metric}
& MAE & RMSE & MAPE\%
& MAE & RMSE & MAPE\%
& MAE & RMSE & MAPE\%
& MAE & RMSE & MAPE\% \\
\midrule

GWNET
& \underline{14.94} & \underline{26.06} & 15.52
& 19.23 & 30.76 & 14.04
& 20.63 & 33.26 & 9.14
& 15.06 & 23.79 & 9.73 \\

DCRNN
& 16.67 & 29.06 & 16.60
& 21.68 & 34.19 & 15.05
& 23.13 & 36.13 & 10.18
& 16.81 & 25.91 & 11.71 \\

AGCRN
& 16.21 & 28.57 & \underline{15.35}
& 19.85 & 32.96 & 13.38
& 21.35 & 35.12 & 9.02
& 16.54 & 26.23 & 10.47 \\

STSGCN
& 18.24 & 30.51 & 17.37
& 21.46 & 34.07 & 14.37
& 24.07 & 38.69 & 10.17
& 17.05 & 26.47 & 10.87 \\

STAEformer
& 16.67 & 29.56 & 17.21
& 18.63 & 30.86 & 13.10
& 19.91 & 33.94 & 8.58
& 13.82 & 23.55 & 9.23 \\

DSTAGNN
& 15.99 & 28.20 & 15.45
& 19.85 & 31.91 & 13.42
& 21.84 & 34.95 & 9.58
& 16.19 & 25.44 & 10.61 \\

STWave
& 15.91 & 28.06 & 16.62
& 19.63 & 32.02 & 13.66
& 21.32 & 35.68 & 8.99
& 14.01 & 23.94 & 9.33 \\

STIDGCN
& 15.98 & 26.66 & 17.24
& \underline{18.38} & \underline{30.47} & \underline{12.42}
& \underline{19.20} & 32.87 & 8.18
& 13.71 & 23.33 & \underline{9.00} \\

DTRformer
& 15.83 & 27.61 & 16.31
& 18.49 & 30.62 & 13.08
& 19.52 & \underline{32.53} & 8.38
& \underline{13.50} & \underline{22.90} & 9.05 \\

LHMformer
& 16.22 & 27.81 & 16.76
& 18.45 & 30.70 & 13.00
& 19.27 & 32.54 & \underline{8.11}
& 13.66 & 23.04 & 9.02 \\

ASTGCN
& 20.25 & 35.99 & 20.67
& 22.20 & 35.24 & 15.66
& 25.50 & 39.75 & 11.43
& 17.07 & 26.49 & 11.80 \\

\midrule
\textbf{SIMBAD}
& \textbf{14.77} & \textbf{25.87} & \textbf{14.35}
& \textbf{18.03} & \textbf{30.08} & \textbf{12.36}
& \textbf{17.86} & \textbf{30.99} & \textbf{7.44}
& \textbf{12.58} & \textbf{22.13} & \textbf{8.35} \\

w/o \(\mathcal{M}\)
& 15.94 & 28.42 & 16.65
& 18.48 & 30.74 & 12.96
& 18.46 & 31.80 & 7.91
& 13.10 & 24.21 & 9.07 \\

w/o \(\mathcal{L}_{u}\)
& 15.19 & 26.99 & 15.25
& 18.25 & 30.32 & 12.60
& 18.23 & 31.40 & 7.69
& 12.91 & 22.31 & 8.79 \\

w/o \(\mathbf{A}\)
& 15.65 & 27.79 & 16.56
& 18.26 & 30.14 & 12.91
& 18.02 & 31.05 & 7.63
& 12.72 & 22.14 & 8.70 \\

\bottomrule
\end{tabular}
}
\end{table*}

Table \ref{tab:overall_ablation} shows the results of eleven baselines, \algshort{} and ablation results. \algshort{} outperforms baseline models in most cases, demonstrating its overall prediction capability. Notably, \algshort{} demonstrates superior performance not only on small-scale datasets such as PEMS08, but also on PEMS07 with large and complex network structures. Specifically, \algshort{} achieved an average performance gain of 5.80\% and 6.66\% on PEMS08 and PEMS07, respectively, over the strongest baseline.

We conduct a series of ablation experiments on SIMBAD. (1) \textit{w/o} $\mathcal{M}$: remove the masking matrix on the input, using every historical data. (2) \textit{w/o} $\mathcal{L}_{u}$: remove uncertainty and weight historical input based on Eq.~\ref{input_target_sim}. (3) \textit{w/o} \textbf{A}: use unweighted adjacency matrices for all input. Table \ref{tab:overall_ablation} shows the effectiveness of different components in our \algshort{} model. For all datasets, removing the masking matrix $M$, which regulates historical inputs based on similarity to current traffic states, leads to the largest performance degradation, highlighting the crucial role of regulating historical inputs. Also, a significant drop in performance is observed by the removal of dynamic weights among nodes (\textit{w/o} \textbf{A}), indicating that regulating dissimilar patterns among connected nodes is also essential for robust spatio-temporal forecasting. Furthermore, the removal of calculating uncertainty $s_{k}$ also causes a substantial decrease in performance, showing the necessity of regularizing misleading signals.

\begin{figure}[t]
    \centering

    \begin{subfigure}[t]{0.48\columnwidth}
        \centering
        \includegraphics[height=4.8cm]{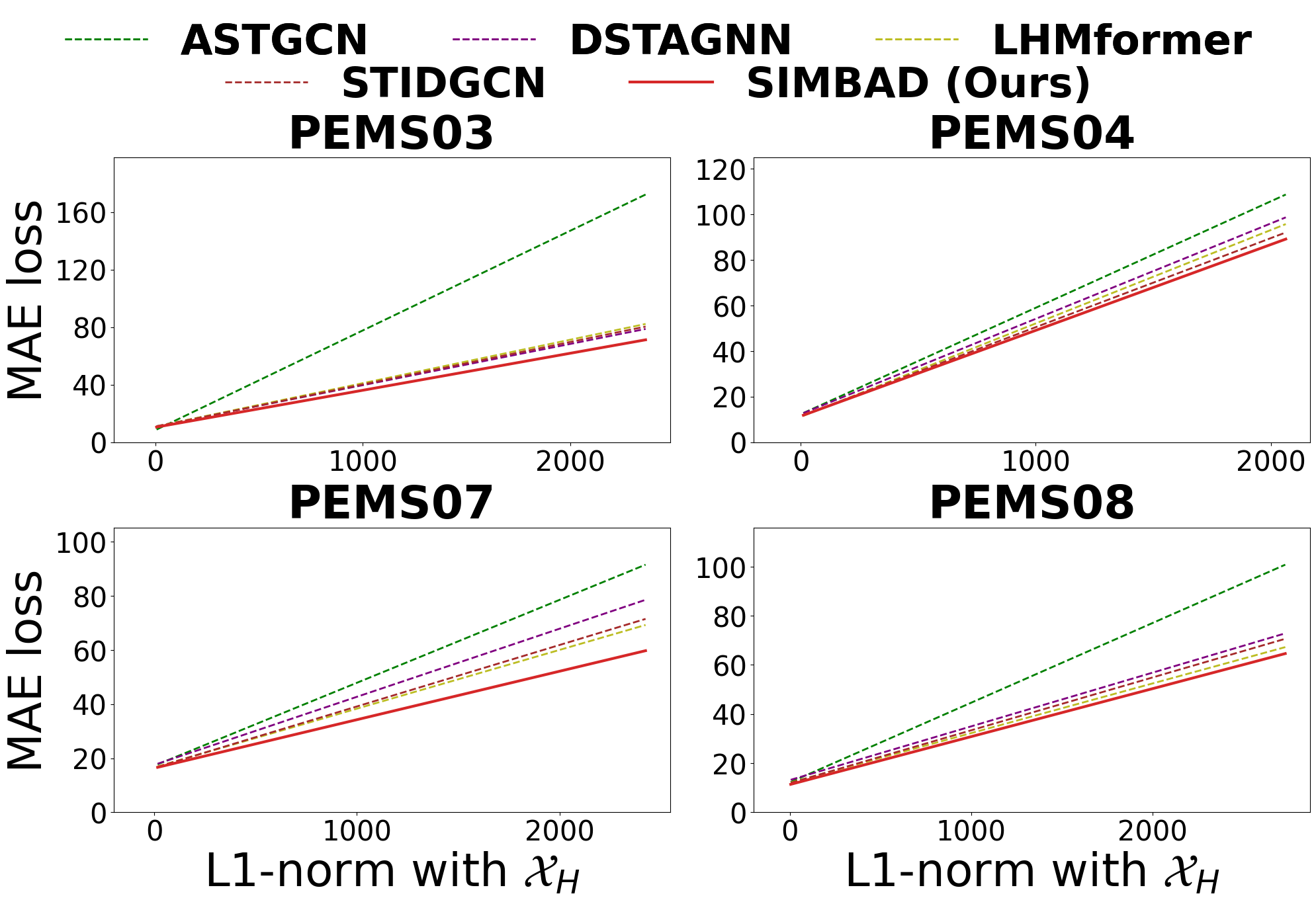}
        \caption{Performance on increasing irregularity.}
    \end{subfigure}
    \begin{subfigure}[t]{0.48\columnwidth}
        \centering
        \includegraphics[height=4.8cm]{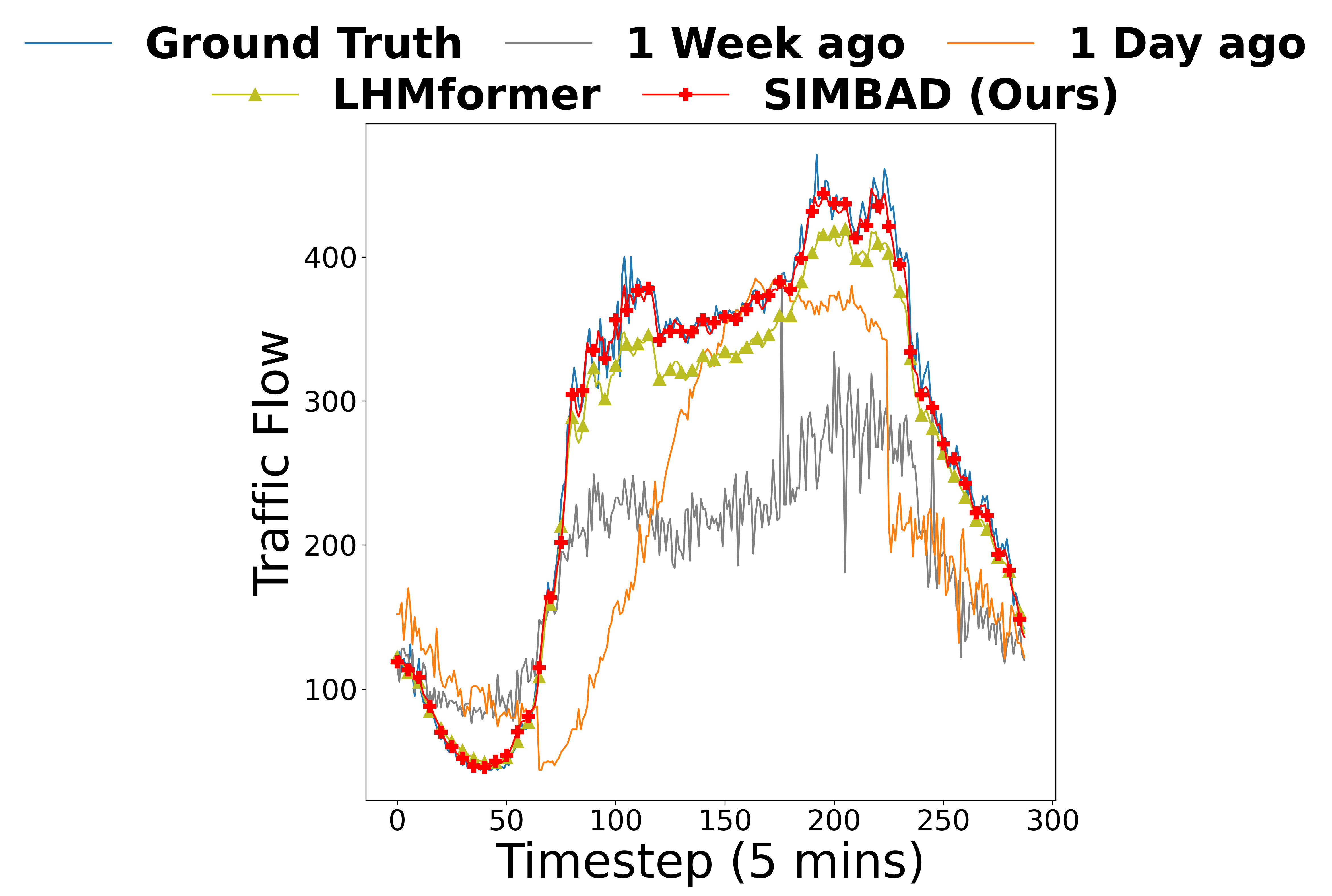}
        \caption{Node 88 on PEMS08.}
    \end{subfigure}

    \caption{
    (a) shows performance comparison over increasing irregularity. 
    (b) shows an example prediction result of \algshort{} and the best recent baseline LHMformer from Node 88 of PEMS08.
    }
    \label{fig_irr}
\end{figure}

\subsection{Performance Analysis with Increasing Irregularity}
    Fig.~\ref{fig_irr}(a) presents the trend of prediction accuracy of all datasets with respect to increasing L1-norm value of target timestep to the past week. We compared SIMBAD to four types of baselines: (1) \textit{ASTGCN}~\cite{astgcn} which leverages historical data for prediction; (2) GNN-based model \textit{DSTAGNN}~\cite{dstagnn}; (3) a Transformer-based \textit{LHMformer}~\cite{lhmformer}; (4) memory-based \textit{STIDGCN}~\cite{stidgcn}. \algshort{} consistently outperforms other baselines as the L1-norm of the target sequence and its corresponding historical data increases, while maintaining the superior overall performance. In particular, the largest performance gain over ASTGCN on all datasets indicates that the regulation of irregular patterns in historical datasets is crucial, especially in forecasting aperiodic patterns. Also, \algshort{} demonstrates its strong robustness to growing aperiodicity throughout all datasets. For instance, in PEMS03, DSTAGNN, LHMformer, STIDGCN exhibit comparable performance across the whole calculated L1-norm range. However, DSTAGNN and LHMformer suffer from substantial performance degradation in PEMS04 and STIDGCN achieves only the third-best performance in PEMS07. These results demonstrate that \algshort{} can effectively distinguish and maintain its robustness to sporadic, irregular patterns while preserving overall performance. 
    
    Fig.~\ref{fig_irr}(b) presents an example of performance comparison between LHMformer~\cite{lhmformer} and \algshort{} under an irregular traffic scenario where the current traffic context significantly differs from both the 1 week and 1 day historical patterns. Since LHMformer~\cite{lhmformer} leverages historical information using a full-day window to emphasize daily periodicity, its prediction becomes strongly influenced by the daily historical pattern, resulting in performance degradation when such patterns fail to generalize to the current context. Additional case study results are provided in Appendix~\ref{ap:irregular}.

\subsection{Hyperparameter Sensitivity Analysis}
\begin{figure*}[t]
    \centering

    \includegraphics[width=0.7\textwidth]{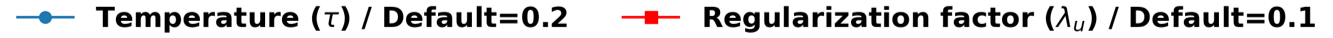}

    \vspace{2mm}

    \begin{subfigure}[t]{0.24\textwidth}
        \centering
        \includegraphics[width=\linewidth]{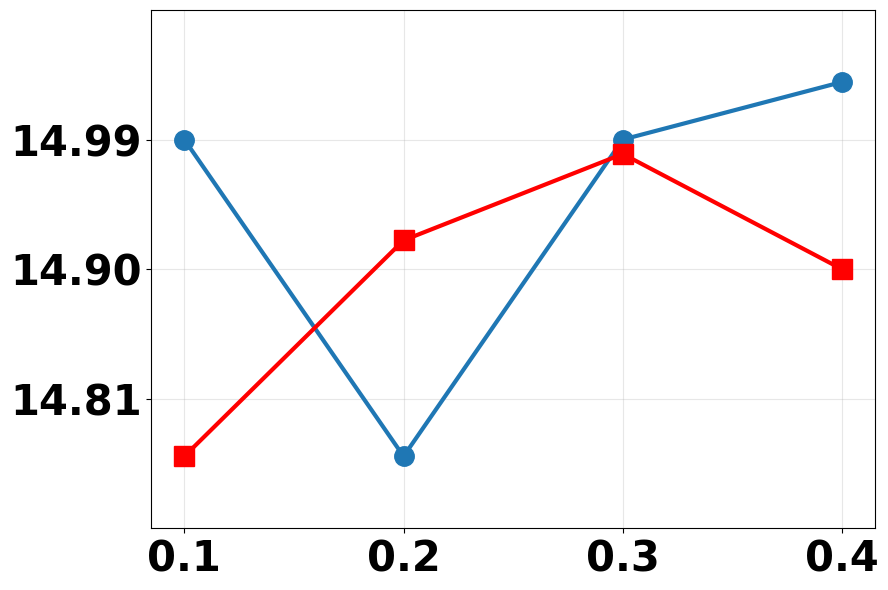}
        \caption{PEMS03.}
        \label{fig:hyper_pems03}
    \end{subfigure}
    \hfill
    \begin{subfigure}[t]{0.24\textwidth}
        \centering
        \includegraphics[width=\linewidth]{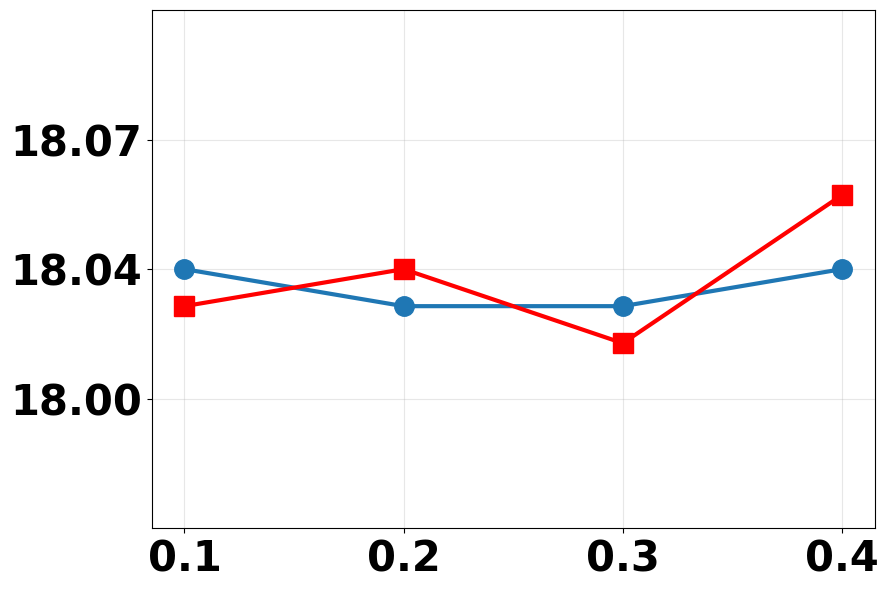}
        \caption{PEMS04.}
        \label{fig:hyper_pems04}
    \end{subfigure}
    \hfill
    \begin{subfigure}[t]{0.24\textwidth}
        \centering
        \includegraphics[width=\linewidth]{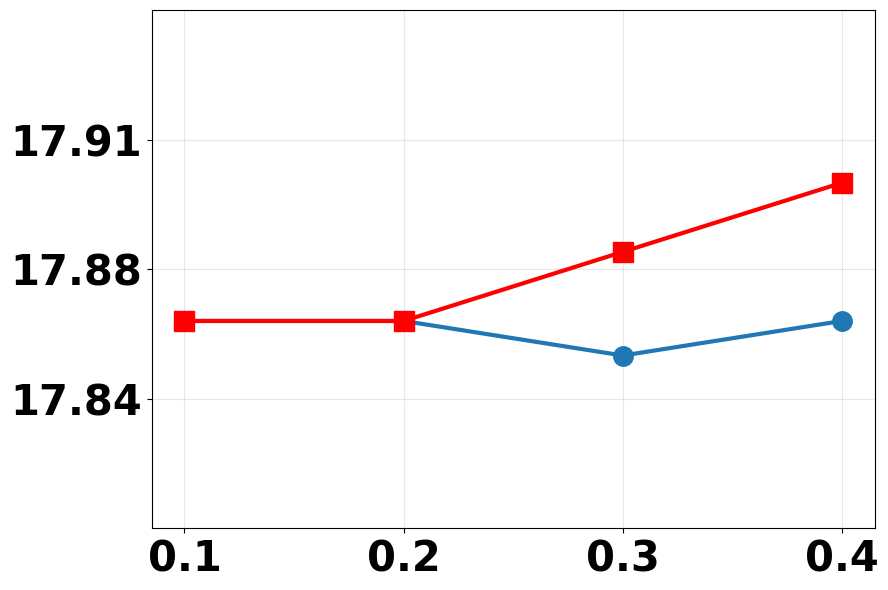}
        \caption{PEMS07.}
        \label{fig:hyper_pems07}
    \end{subfigure}
    \hfill
    \begin{subfigure}[t]{0.24\textwidth}
        \centering
        \includegraphics[width=\linewidth]{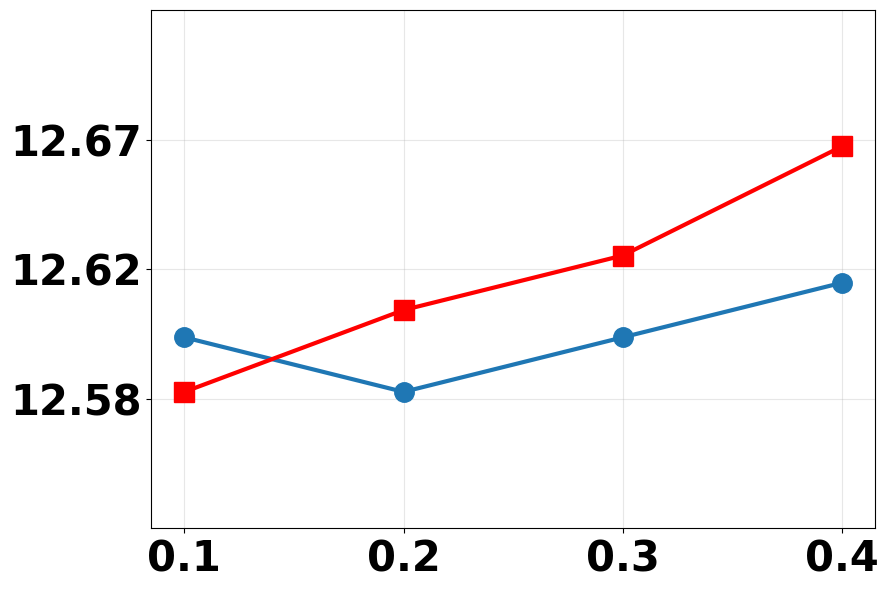}
        \caption{PEMS08.}
        \label{fig:hyper_pems08}
    \end{subfigure}

    \caption{
    Hyperparameter sensitivity analysis in terms of MAE on four datasets.
    }
    \label{fig:hyperparameter}
\end{figure*}
We further investigate hyperparameter sensitivity on the tested datasets. We select two hyperparameters for investigation: (1) temperature $\tau$, which controls the sharpness of the distribution in Eq.~\ref{distribution}; (2) regularization factor $\lambda_u$, which regularizes the loss in Eq.~\ref{unc_loss}. The default values for $\tau$ and $\lambda_u$ are 0.2 and 0.1, respectively, and the results are shown in Fig.~\ref{fig:hyperparameter}. 

Results demonstrate that \algshort{} maintains stable performance across a wide range of hyperparameter configurations. Furthermore, performance degradation can be observed with a high temperature $\tau$ value. A high $\tau$ value leads to a smoother distribution in Eq.~\ref{distribution} and reduces the distinction between reliable and unreliable periodic inputs. This causes the dissimilar historical patterns to be more likely to be aggregated, weakening the similarity-based periodic selection process and leading to performance degradation. Also, as the regularization factor $\lambda_u$ increases, the uncertainty regularization term exerts a stronger influence on the optimization process. Excessive regularization may overemphasize uncertainty minimization rather than forecasting accuracy, which can suppress useful periodic information and degrade the overall prediction performance.

\subsection{Modularity of SIMBAD}

\begin{wraptable}{r}{0.52\columnwidth}
\vspace{-2mm}
\centering
\caption{Modularization results in terms of MAE.}
\label{modularization}
\footnotesize
\setlength{\tabcolsep}{3pt}
\begin{tabular}{lcccc}
\toprule
\textsc{Model} & PEMS03 & PEMS04 & PEMS07 & PEMS08 \\
\midrule
DSTAGNN
& \textbf{15.99} & 19.85 & 21.84 & 16.19 \\
+ Ours
& 16.44 & \textbf{19.53}
& \textbf{21.14} & \textbf{14.89} \\
\cmidrule(lr){1-5}
STIDGCN
& 15.98 & 18.38 & 19.20 & 13.71 \\
+ Ours
& \textbf{15.20} & \textbf{18.27}
& \textbf{18.83} & \textbf{13.12} \\
\cmidrule(lr){1-5}
LHMformer
& 16.22 & \textbf{18.45} & \textbf{19.27} & 13.66 \\
+ Ours
& \textbf{16.15} & 18.83
& 19.76 & \textbf{13.53} \\
\bottomrule
\end{tabular}
\vspace{-2mm}
\end{wraptable}

To evaluate its applicability beyond \algshort{}, we integrate the temporal regulation mechanism in \S 3.1.1 into existing forecasting architectures. The module regulates periodic inputs based on their relevance and uncertainty before passing them to the forecasting backbone, allowing each baseline's original forecasting architectures to be preserved. This modular design enables existing architectures to selectively exploit historical information without redesigning their core components.

As shown in Table~\ref{modularization}, the module reduces MAE in 9 of 12 cases when equipped with three representative baselines. DSTAGNN and STIDGCN achieve average relative reductions of 2.51\% and 2.93\%, respectively, with improvements reaching 8.03\% and 4.30\% on PEMS08. LHMformer achieves reductions of 0.43\% on PEMS03 and 0.95\% on PEMS08. These results demonstrate that regulating historical information according to its relevance and uncertainty can also enhance existing forecasting architecture. This confirms that the proposed temporal regulation mechanism can serve as a general and modular component for improving robustness to irregular historical patterns. Additional results on other baselines and metrics are provided in Appendix~\ref{ap:modularization}.

\subsection{Historical Window Analysis}

\begin{table*}[!htbp]
\centering
\caption{Performance comparison on different historical windows.}
\label{historical}
\vspace{2mm}
\resizebox{\textwidth}{!}{
\begin{tabular}{@{}l ccc ccc ccc ccc@{}}
\toprule
\textsc{Model}
& \multicolumn{3}{c}{PEMS03}
& \multicolumn{3}{c}{PEMS04}
& \multicolumn{3}{c}{PEMS07}
& \multicolumn{3}{c}{PEMS08} \\
\cmidrule(lr){2-4}
\cmidrule(lr){5-7}
\cmidrule(lr){8-10}
\cmidrule(lr){11-13}
\textsc{Metric}
& MAE & RMSE & MAPE\%
& MAE & RMSE & MAPE\%
& MAE & RMSE & MAPE\%
& MAE & RMSE & MAPE\% \\
\midrule

\textbf{SIMBAD}
& \textbf{14.77} & \textbf{25.87} & \textbf{14.35}
& \textbf{18.03} & \textbf{30.08} & \textbf{12.36}
& \textbf{17.86} & \textbf{30.99} & \textbf{7.44}
& \textbf{12.58} & \textbf{22.13} & \textbf{8.35} \\

\midrule

No week
& 15.29 & 26.79 & 16.14
& 19.54 & 31.59 & 13.19
& 19.56 & 32.93 & 8.32
& 14.55 & 23.52 & 9.41 \\

No day
& 15.02 & 25.91 & 14.55
& 18.30 & 30.40 & 12.56
& 18.09 & 31.35 & 7.50
& 12.95 & 22.53 & 8.68 \\

\bottomrule
\end{tabular}
}
\end{table*}
In this section, we analyze the contribution of each historical window. In our settings, we used 2 weeks, 1 week and 1 day periodic data for prediction. We show results when the weekly historical data and daily historical data are not considered for calculation, respectively. Results are shown in Table~\ref{historical}. As shown in the table, the historical information is indeed important for accurate prediction. In particular, removing the weekly historical data leads to an average MAE increase of 9.27\% across all datasets, compared to only 1.86\% when the daily historical data is removed. The most pronounced degradation is observed on PEMS08, where removing the weekly historical data increases MAE by 15.66\%, compared to 2.94\% when the daily historical data is removed. These results suggest that weekly historical information plays a more critical role in accurate traffic prediction. This is a valid result, considering that traffic data exhibits strong weekly periodicity, where mobility patterns are largely governed by recurring schedules such as weekday commuting and weekend behavior.





\section{Related Works}

Traditional time series forecasting methods, such as ARIMA~\cite{arima}, VAR~\cite{var} and SVR~\cite{svm}, focused on modeling temporal dependencies, and recurrent architectures, such as RNNs and LSTMs, were introduced to better capture temporal context. Then, traffic forecasting studies~\cite{stgcn, astgcn, stsgcn} incorporated spatial modeling through graph-based approaches, enabling representation of interdependencies among road sensors and capturing spatial information.

\spara{Spatio-Temporal Traffic Forecasting.} Based on spatial relations between sensors, traffic forecasting requires handling spatial correlations between data, different from general time series prediction.
To effectively model the traffic flow based on both spatial and temporal dependencies, various studies have been conducted.
DCRNN~\cite{dcrnn} models the traffic flow as a diffusion process, STGCN~\cite{stgcn} incorporates 1D temporal convolution along with spatial graph convolution to capture spatio-temporal dependencies.
Models such as ASTGCN~\cite{astgcn}, STAEformer~\cite{stae}, LHMformer~\cite{lhmformer} incorporated attention mechanisms which achieved promising performance.
PDFormer~\cite{pdformer}, GMAN~\cite{gman}, STID~\cite{stid} applied spatial embedding to capture inherent spatial dependencies.
To capture the temporal and spatial dependencies simultaneously, STSGCN~\cite{stsgcn} modeled the spatial and temporal blocks altogether.
STFGNN~\cite{stfgnn} and PDFormer~\cite{pdformer} implemented DTW~\cite{dtw} to measure the similarity between nodes and used them to build a similarity-based temporal graph. Recent models, such as PatchSTG~\cite{patch}, achieved great performance by combining Transformer with a tree structure, but it requires auxiliary information such as latitude and longitude.

\spara{Memory-Based Traffic Forecasting.} With the success of memory networks in natural language processing~\cite{memory}, various traffic forecasting studies also adopted the idea of memory networks to store and utilize repeating traffic patterns. For instance, PM-MemNet~\cite{pmmemnet} proposes that traffic data can be summarized into numerous clusters of representative patterns by using pre-calculated patterns. It matches input data to representative patterns with a key-value structure and uses the gained information from the memory for traffic forecasting. MMNCCRN~\cite{mmnccrn} combines the concepts of memory network and cluster to capture low- and high-level patterns more accurately. Also, MegaCRN~\cite{megacrn} leverages the idea of memory networks by building a Meta-Node Bank. Based on this bank, MegaCRN generates an adaptive meta-graph that is robust and adaptive to normal and non-stationary traffic situations. These memory-based approaches require the employment of a dedicated memory component for exploiting traffic patterns in addition to the traffic forecasting model, whereas \algshort{} exploits only the similarity-based proxy derived from the input sequences to adaptively control the periodic traffic patterns.
\smallskip
\section{Conclusion}
In this work, we introduce \algshort{}, a novel spatio-temporal traffic forecasting framework to enhance prediction performance, particularly under aperiodicity. By adaptively balancing the influence of periodic temporal patterns and spatial dependencies through similarity-aware weighting, \algshort{} addresses the limitations of existing approaches when confronted with unexpected traffic fluctuations observed in historical periodic inputs. \algshort{} exploits simple and lightweight similarity-based weights through an aperiodicity-aware modulation layer and a spatio-temporal modeling layer to refine the flow of temporal information across graph nodes, leading to more robust and accurate forecasting. Comprehensive experiments on seven real-world datasets validate the effectiveness of \algshort{}, demonstrating its superior performance in both regular and irregular traffic scenarios. Furthermore, the modularity of \algshort{} has been demonstrated by showing that adopting the temporal regulation mechanism in existing baselines can improve their performance. These results underline the potential of exploiting simple similarity-based historical signals in advancing Intelligent Transportation Systems and supporting more reliable traffic management.

\begin{ack}
This work was supported by the Institute of Information \& Communications Technology Planning \& Evaluation (IITP) grants (No. IITP-2026-RS-2025-02304828(40\%), No. IITP-2026-RS-2024-00436857(20\%), No. RS-2026-25507543(20\%), No. IITP-2026-RS-2020-II201819(10\%)) and the National Research Foundation of Korea (NRF) grant (No. RS-2025-25435830(10\%)) funded by the Korea government (MSIT).
\end{ack}

\bibliographystyle{plainnat}
\bibliography{ref}

\newpage
\appendix

\section{Case Study}
\subsection{Effect of Uncertainty}
In this section, we further analyze the effect of uncertainty $s_{k}$ we proposed. The uncertainty $s_{k}$ reflects the amount of discrepancy between $d_{k}^{-}$ and $d_{k}^{+}$ in Eq.~\ref{input_target_sim}, where high $s_{k}$ value means high discrepancy. Therefore, as in Eq.~\ref{distribution}, we weight the predicted similarity to reduce the contribution of historical instances whose input context conflicts with the target. To validate this behavior, we visualize the effect of uncertainty on datasets PEMS04 and PEMS08 to observe whether uncertainty properly reduces the contribution of historical instances with high discrepancy.

\begin{figure}[H]
\centering
\begin{subfigure}{1.00\textwidth}
\centering
\includegraphics[width=0.9\linewidth]{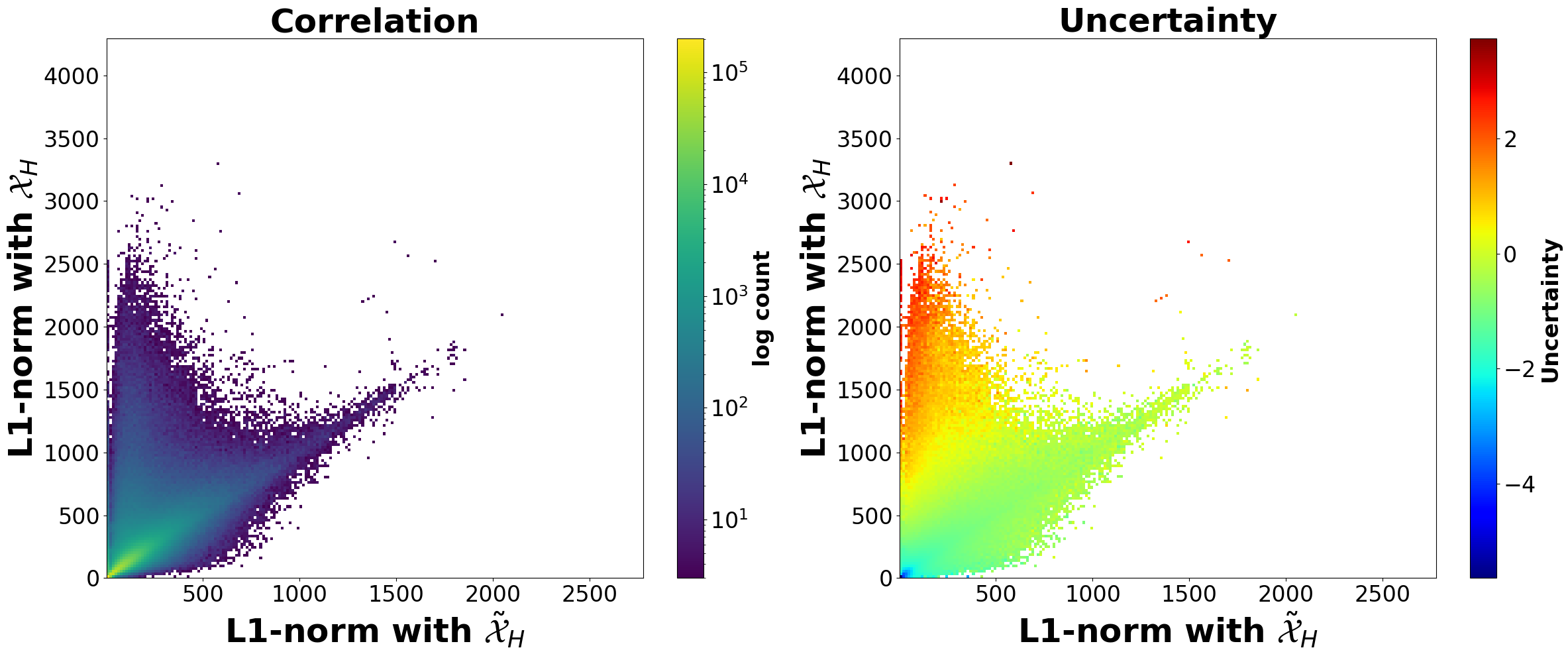}
\caption{PEMS04.}
\end{subfigure}
\begin{subfigure}{1.00\textwidth}
\centering
\includegraphics[width=0.9\linewidth]{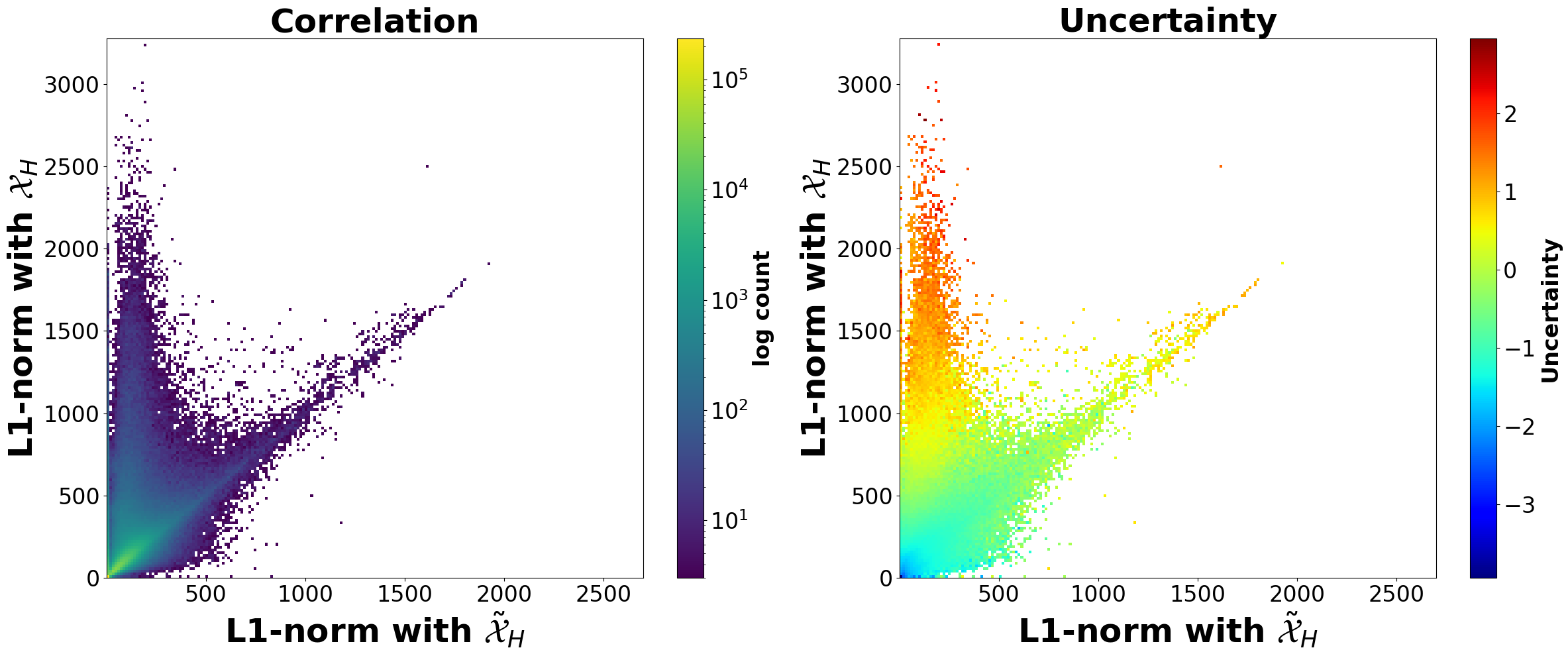}
\caption{PEMS08.}
\end{subfigure}
\caption{For each dataset, the left figure shows correlation between L1-norm of target and aggregated historical input $\mathcal{X}_{H}$ in Eq.~\ref{aggregated} to L1-norm of target and the periodic inputs in Eq.~\ref{periodic_input}. The right figure shows the uncertainty value for each L1-norm correlation. Higher value means higher uncertainty.}
\label{fig_uncertainty}
\end{figure}

The left-side of Fig.~\ref{fig_uncertainty} shows that correlation of L1-norm is mostly positive, implying that the aggregated historical input $\mathcal{X}_{H}$ generally follows the traffic flow tendency to that of the target. However, there exist sections where the L1-norm with $\tilde{\mathcal{X}}_{H}$ is low, whereas the L1-norm with $\mathcal{X}_{H}$ is high. As illustrated in the right panels, uncertainty assigns the mismatched sections as \emph{uncertain}, thus reducing the contribution of the unreliable inputs as in Eq.~\ref{distribution}. This demonstrates that the proposed uncertainty mechanism successfully identifies and suppresses the misleading historical patterns.

Also, there is an extra interesting observation that the lower-right region of the plots, where the L1-norm with $\mathcal{X}_H$ is high but the L1-norm with the target is low, is largely absent. Notably, as shown in Fig.~\ref{fig1b}, data points in such region are observed. However, our aggregated input does not exhibit this phenomenon, demonstrating that our model does not mistakenly mask the highly-correlated situations.

\subsection{Irregular Pattern Prediction}
\label{ap:irregular}

\begin{figure}[H]
\centering
\begin{subfigure}{1.00\textwidth}
\centering
\includegraphics[width=0.95\linewidth]{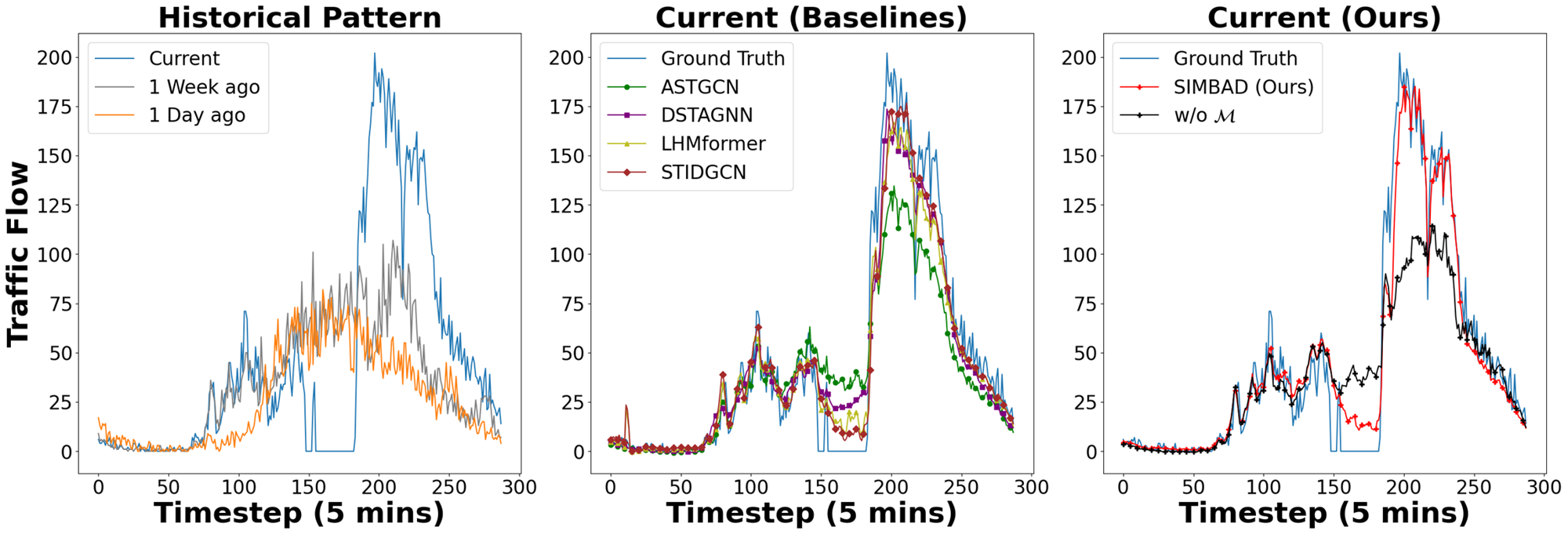}
\caption{PEMS03 -- Node 209.}
\label{fig_appendix_irregular_a}
\end{subfigure}
\begin{subfigure}{1.00\textwidth}
\centering
\includegraphics[width=0.95\linewidth]{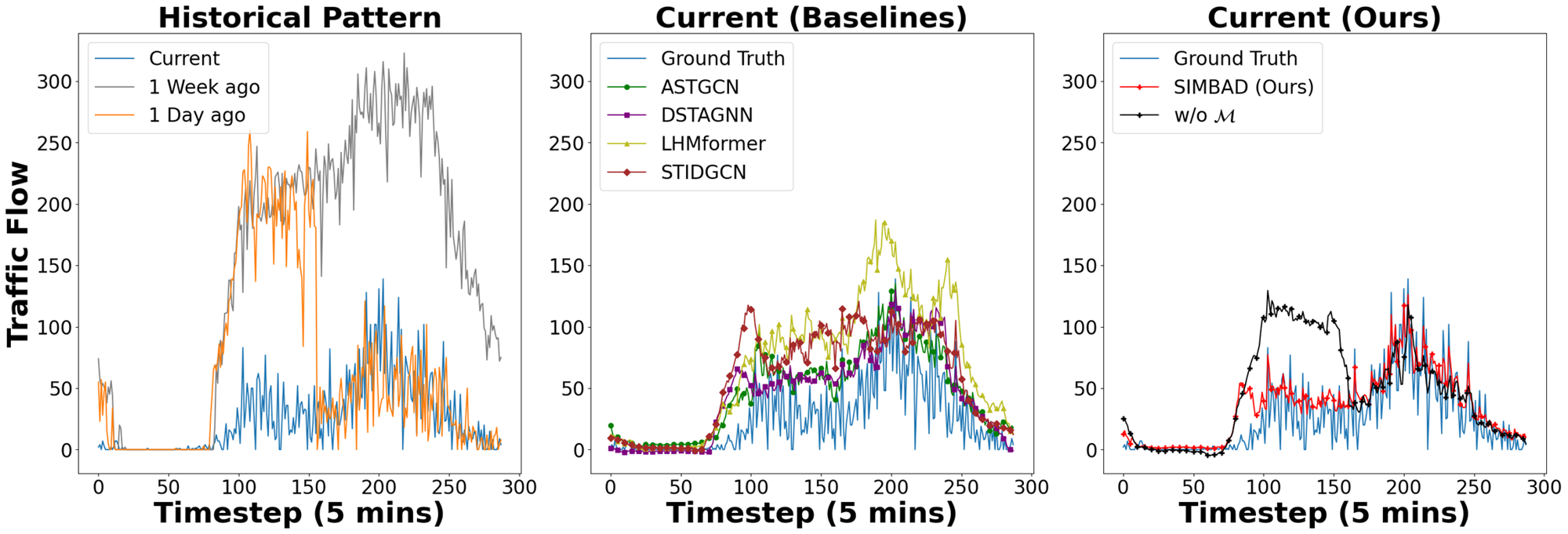}
\caption{PEMS04 -- Node 247.}
\label{fig_appendix_irregular_b}
\end{subfigure}
\begin{subfigure}{1.00\textwidth}
\centering
\includegraphics[width=0.95\linewidth]{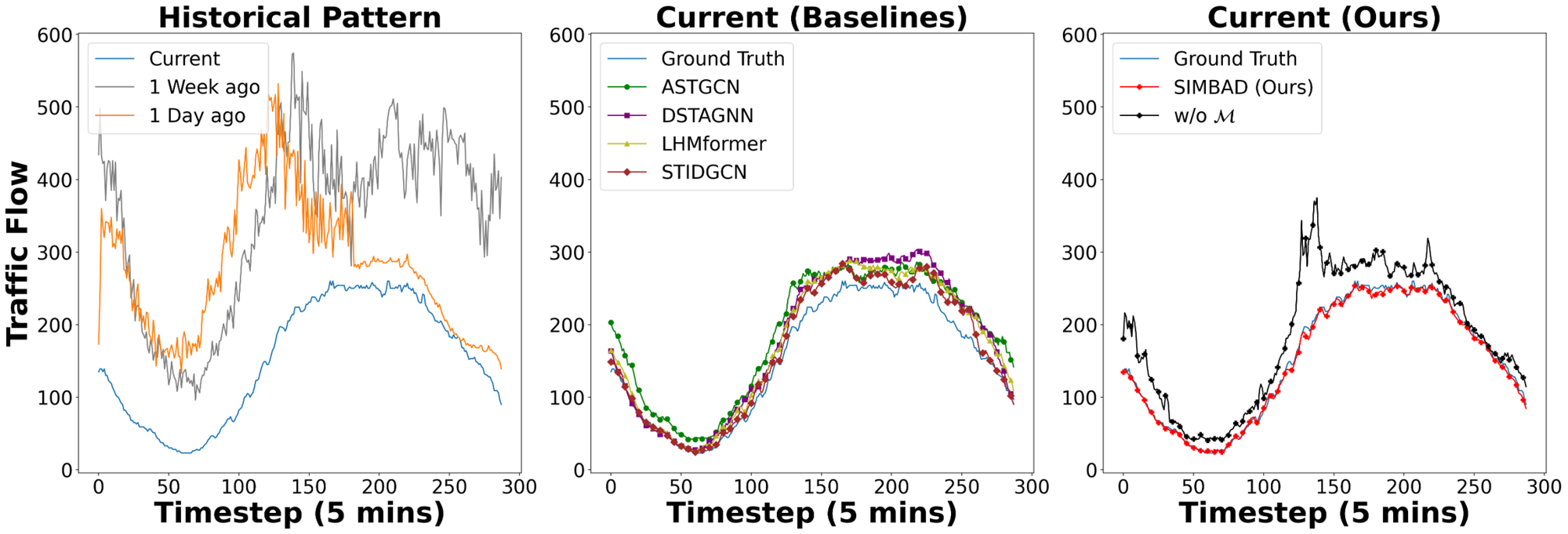}
\caption{PEMS07 -- Node 674.}
\label{fig_appendix_irregular_c}
\end{subfigure}
\begin{subfigure}{1.00\textwidth}
\centering
\includegraphics[width=0.95\linewidth]{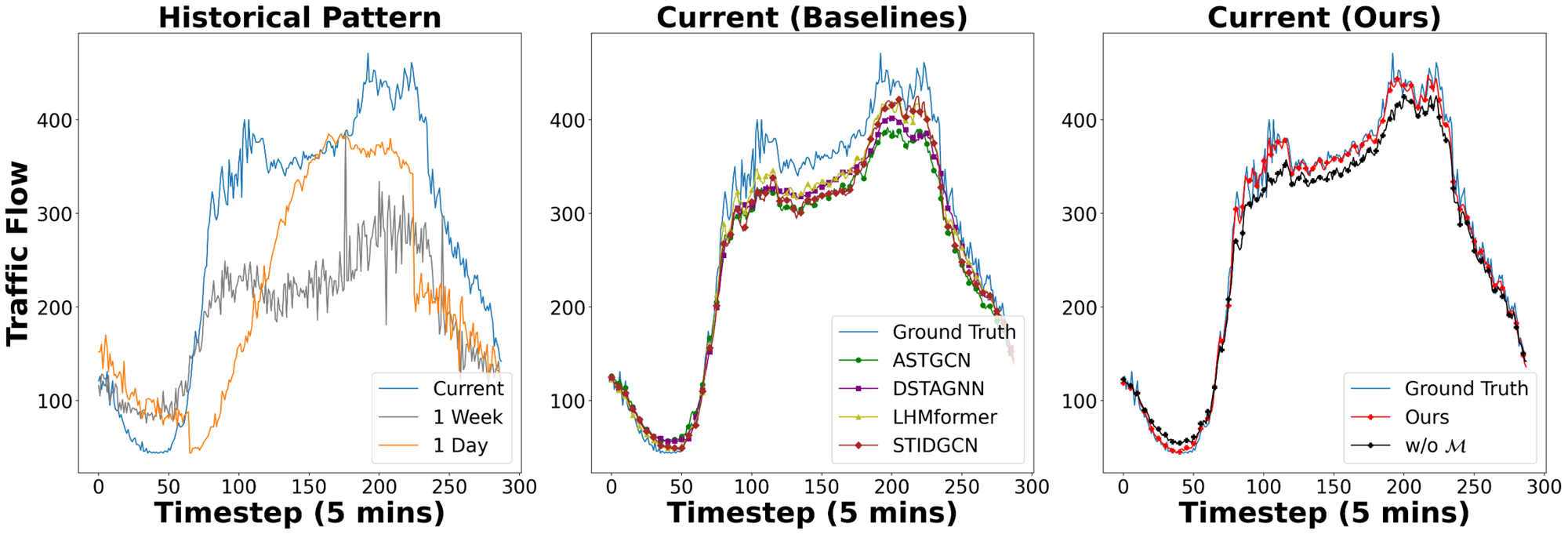}
\caption{PEMS08 -- Node 88.}
\label{fig_appendix_irregular_d}
\end{subfigure}
\caption{Left-side figure shows the current and the respective daily and weekly historical trends. Middle figure shows the prediction performance of comparing baselines. Right-side figure shows the prediction performance of \algshort{} and \algshort{} without masking matrix $\mathcal{M}$ of Eq.~\ref{mask}.}
\label{fig_appendix_irregular}
\end{figure}

In this section, we evaluate the model performance on irregular timesteps, where the target traffic patterns deviate significantly from historical trends. We mainly focus on representative cases where both weekly and daily patterns differ from the target, as traffic data generally follows strong temporal periodicity, where the same day of the week shows similar patterns and recent daily observations capture daily recurring trends.

As illustrated in Fig.~\ref{fig_appendix_irregular}, baseline models inherit dominant historical trends, leading to systematic errors when these patterns fail to generalize to the current context. For example, ASTGCN~\cite{astgcn} directly utilizes weekly and daily historical inputs without considering their reliability, leading to incorrect prediction of the evening peak. As shown in Fig.~\ref{fig_appendix_irregular}(a), the discrepancy between the weekly and daily evening peak patterns introduces conflicting signals, which degrades the prediction accuracy. Also, LHMformer~\cite{lhmformer} leverages the historical data with a full-day window, which emphasizes recent daily patterns. As a result, as shown in Fig.~\ref{fig_appendix_irregular}(d), the predicted magnitude of the evening peak tends to follow that of the previous day.

In contrast, our model addresses these limitations through an uncertainty-aware aggregation mechanism, which evaluates both the relevance and reliability of historical inputs. This allows the model to suppress misleading patterns and accurately capture the true traffic dynamics, even when both weekly and daily references are inconsistent.

We also evaluate the influence of masking matrix $\mathcal{M}$ in Eq.~\ref{mask}, which filters out dissimilar labels based on the calculated similarity between historical data. Results show that the masking matrix is critical for correctly predicting the irregular traffic patterns. For example, in Fig.~\ref{fig_appendix_irregular}(b), \algshort{} without masking matrix tends to predict the morning peaks to follow as in previous day and week. Also, in Fig.~\ref{fig_appendix_irregular}(c), \algshort{} without mask also predicts the morning peaks similar to the previous day.

\newpage

\subsection{Detection of Irregularity}
\label{app:internal}
\begin{figure}[H]
    \centering

    \includegraphics[width=0.75\textwidth]{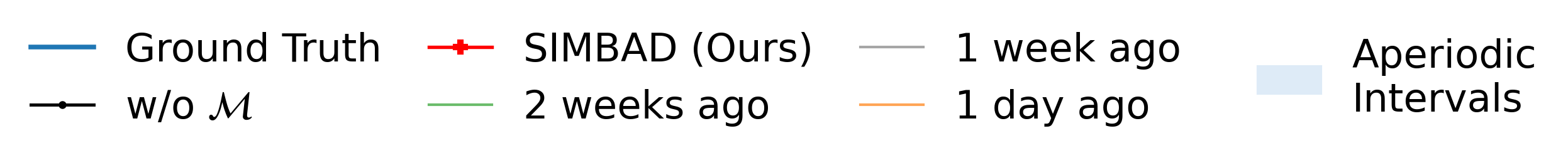}

    \vspace{2mm}

    \begin{subfigure}[t]{0.48\textwidth}
        \centering

        \includegraphics[width=\linewidth]{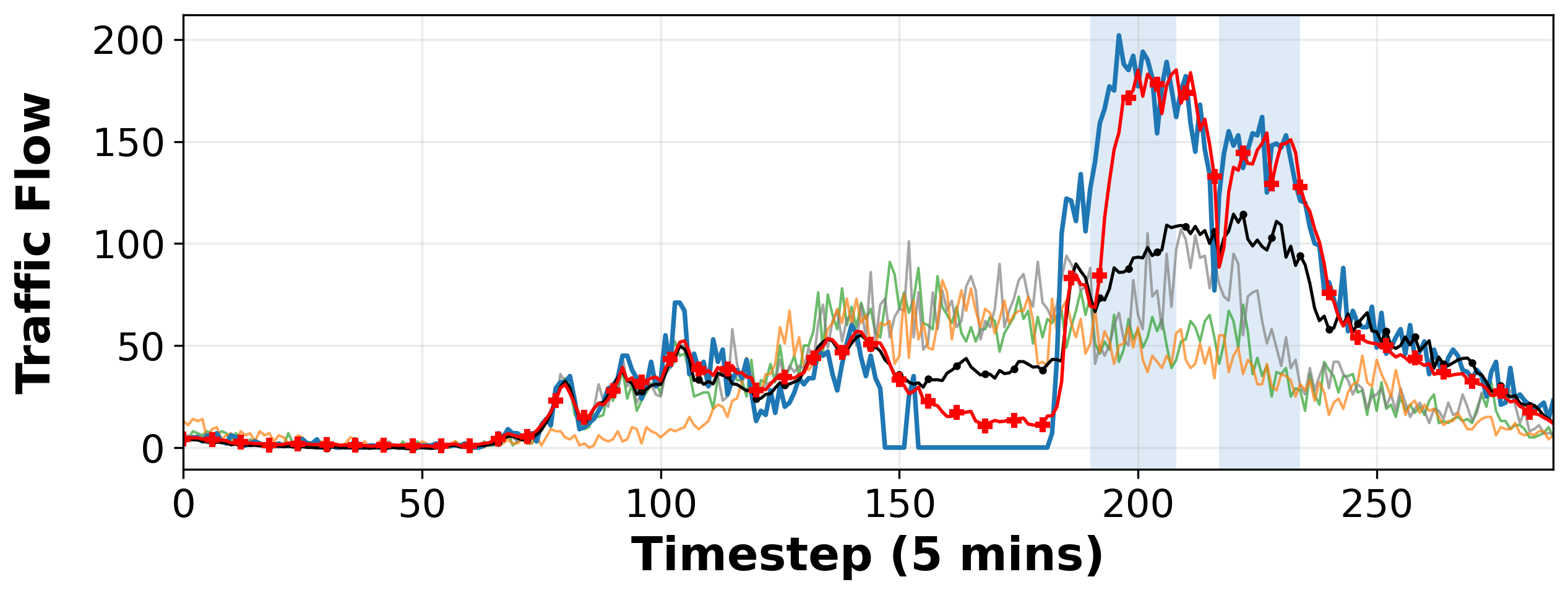}\\[-1mm]

        \includegraphics[width=\linewidth]{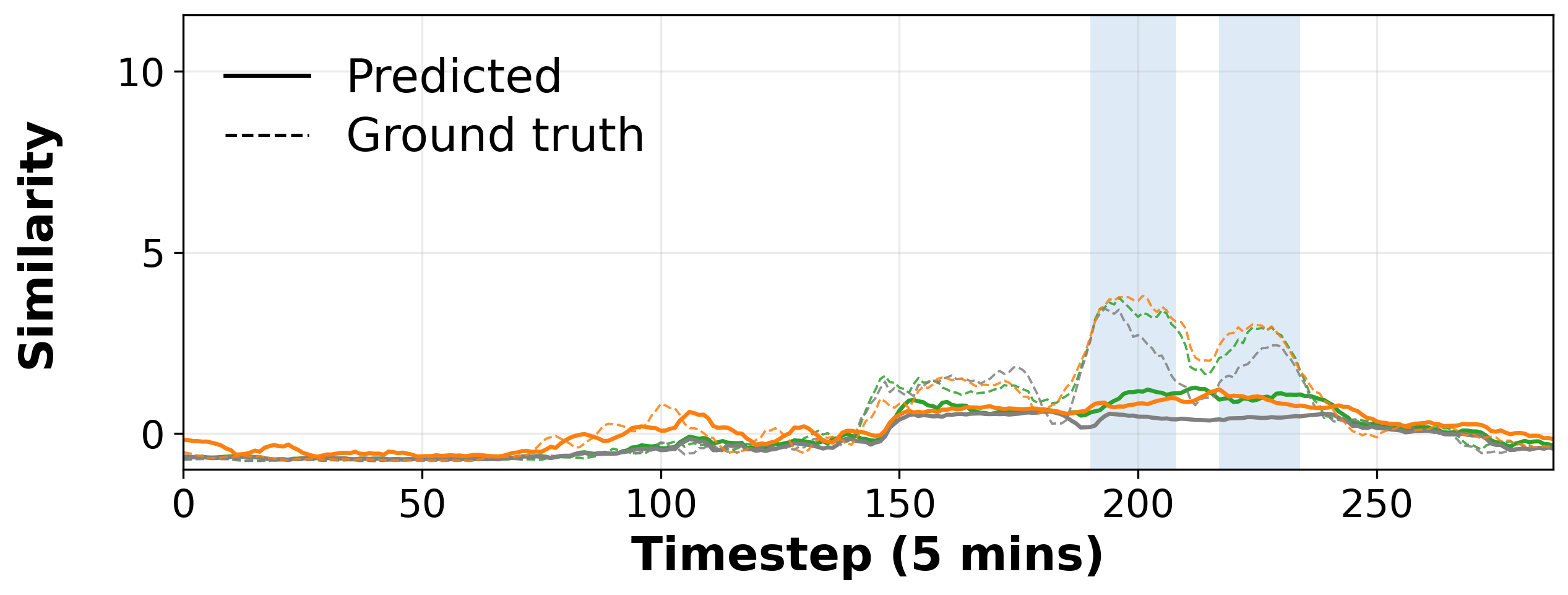}\\[-1mm]

        \includegraphics[width=\linewidth]{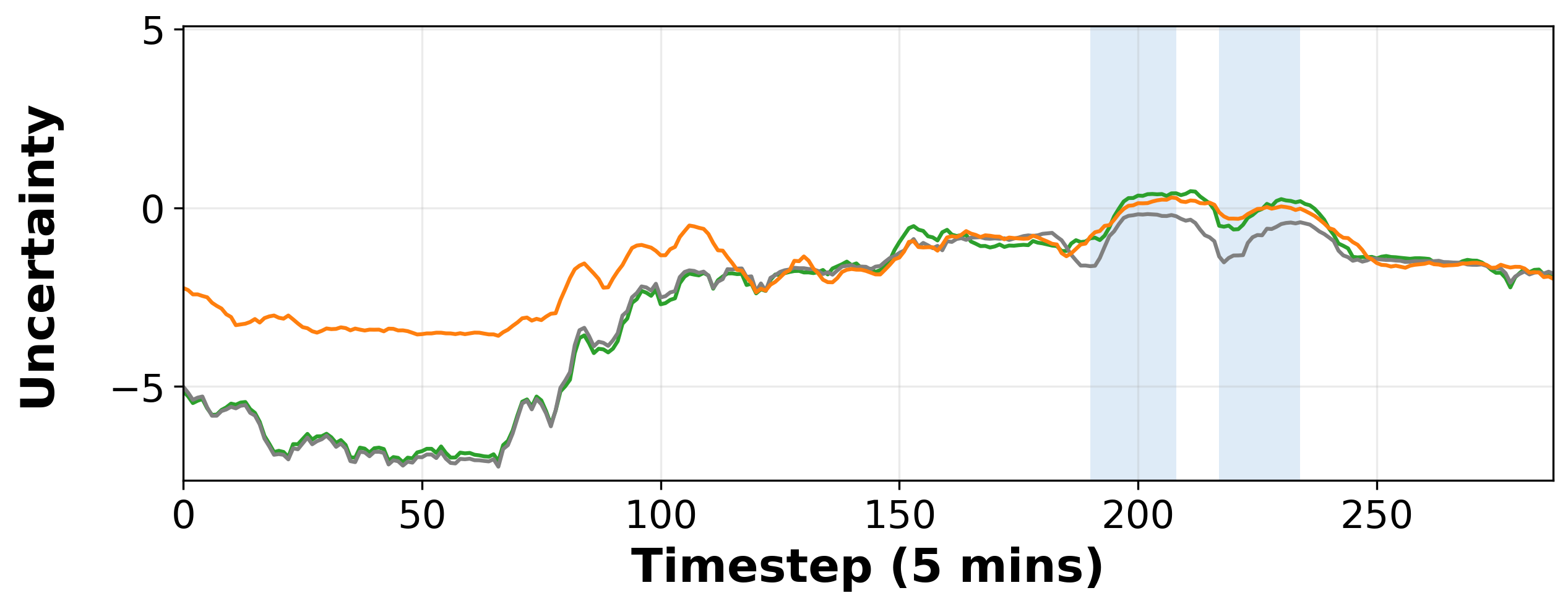}\\[-1mm]

        \includegraphics[width=\linewidth]{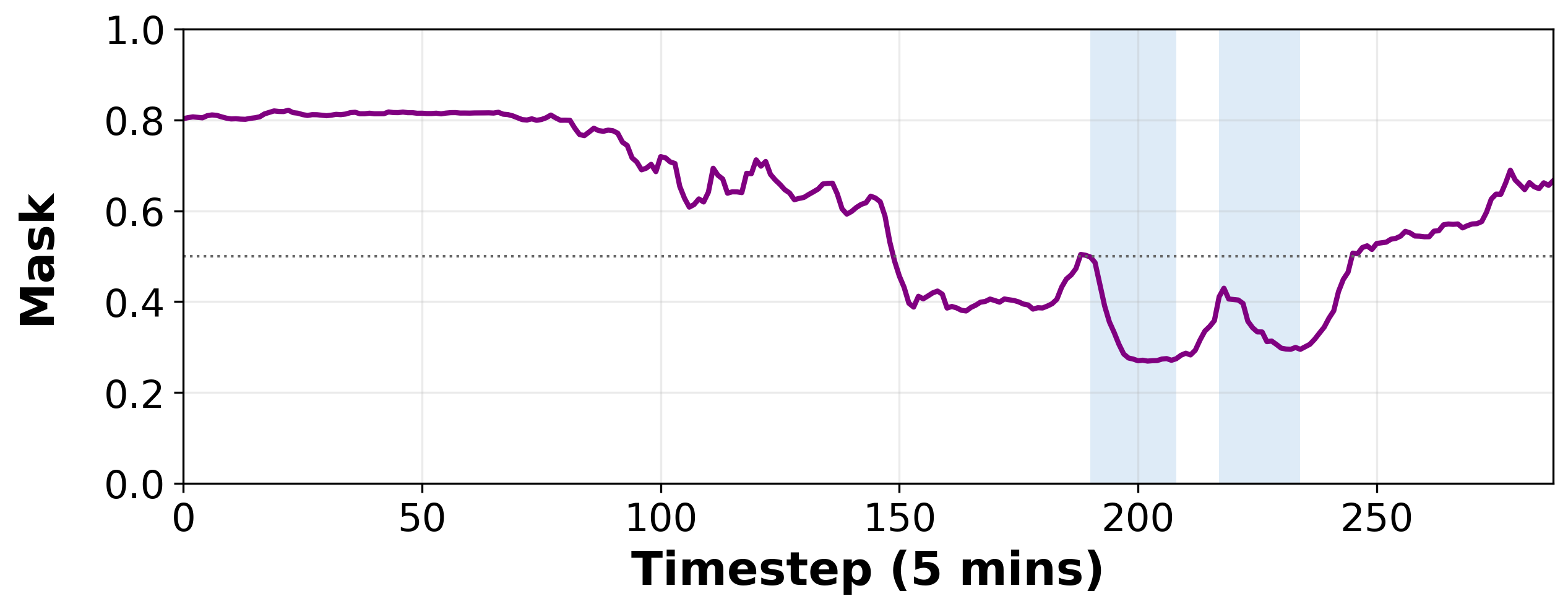}\\[-1mm]

        \includegraphics[width=\linewidth]{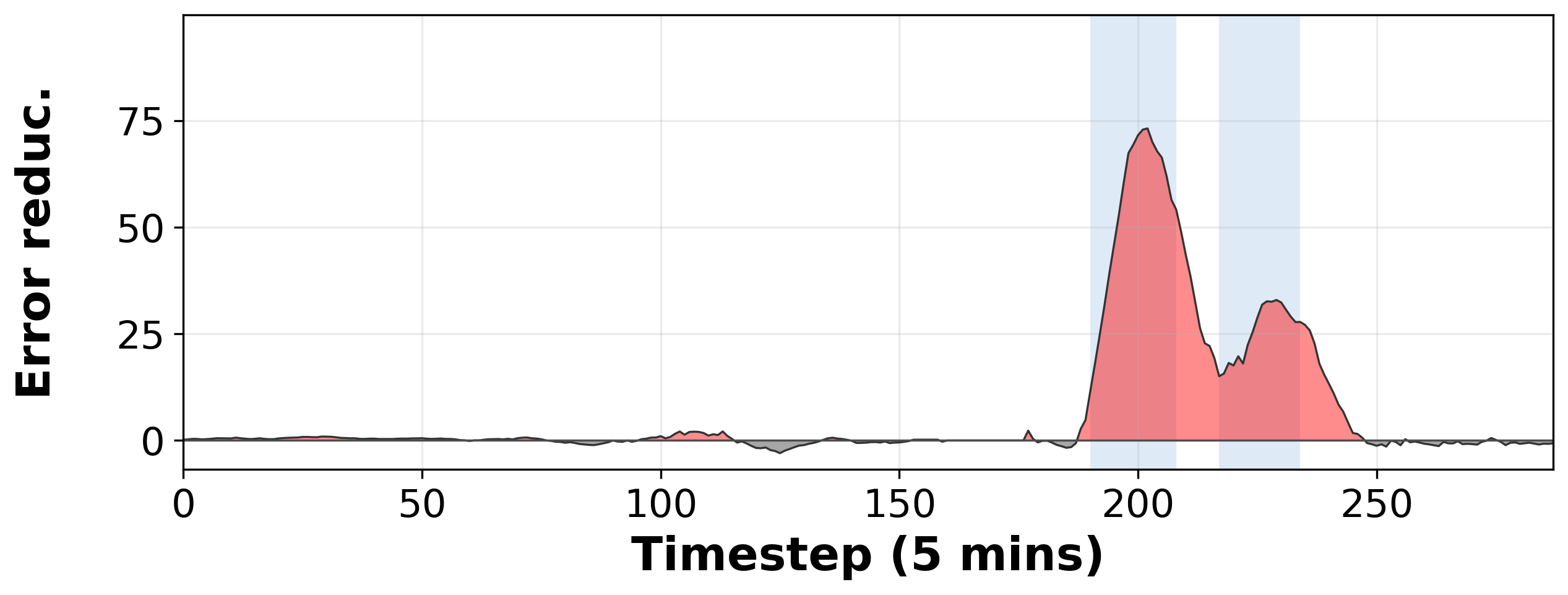}

        \caption{PEMS03 -- Node 209.}
        \label{fig:pems03-node209}
    \end{subfigure}
    \hfill
    \begin{subfigure}[t]{0.48\textwidth}
        \centering

        \includegraphics[width=\linewidth]{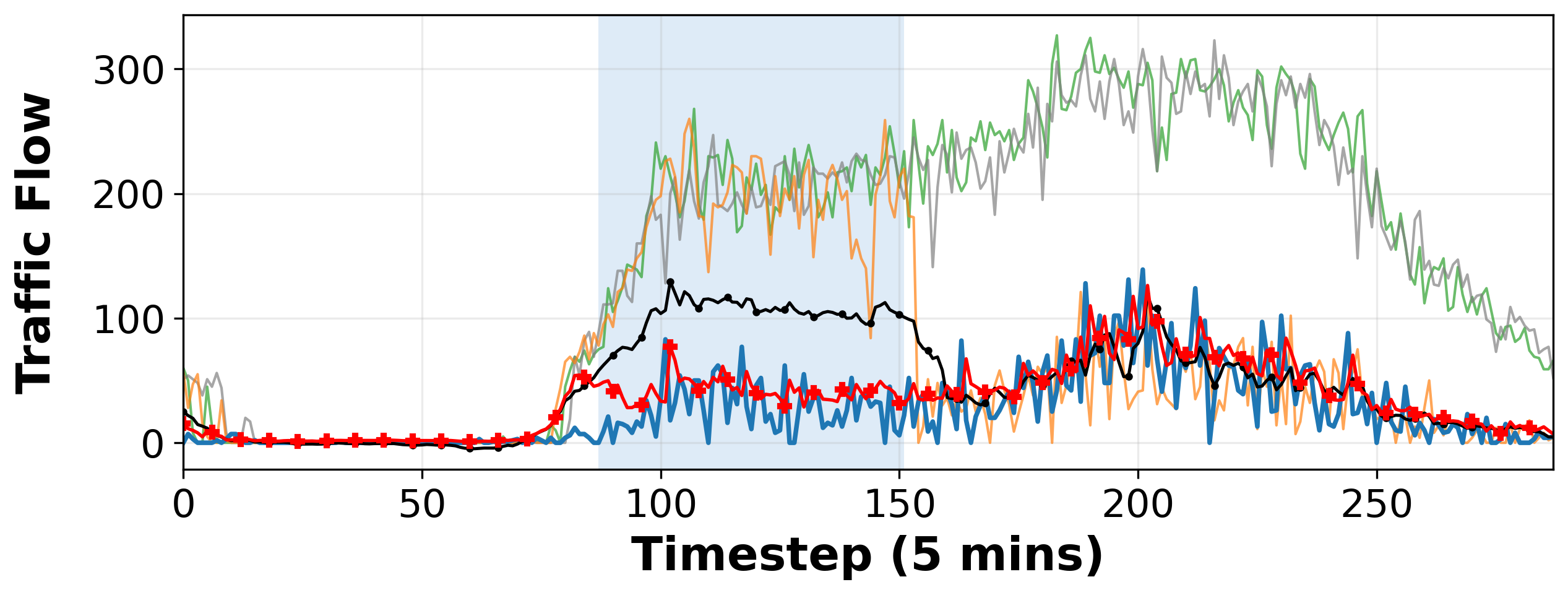}\\[-1mm]

        \includegraphics[width=\linewidth]{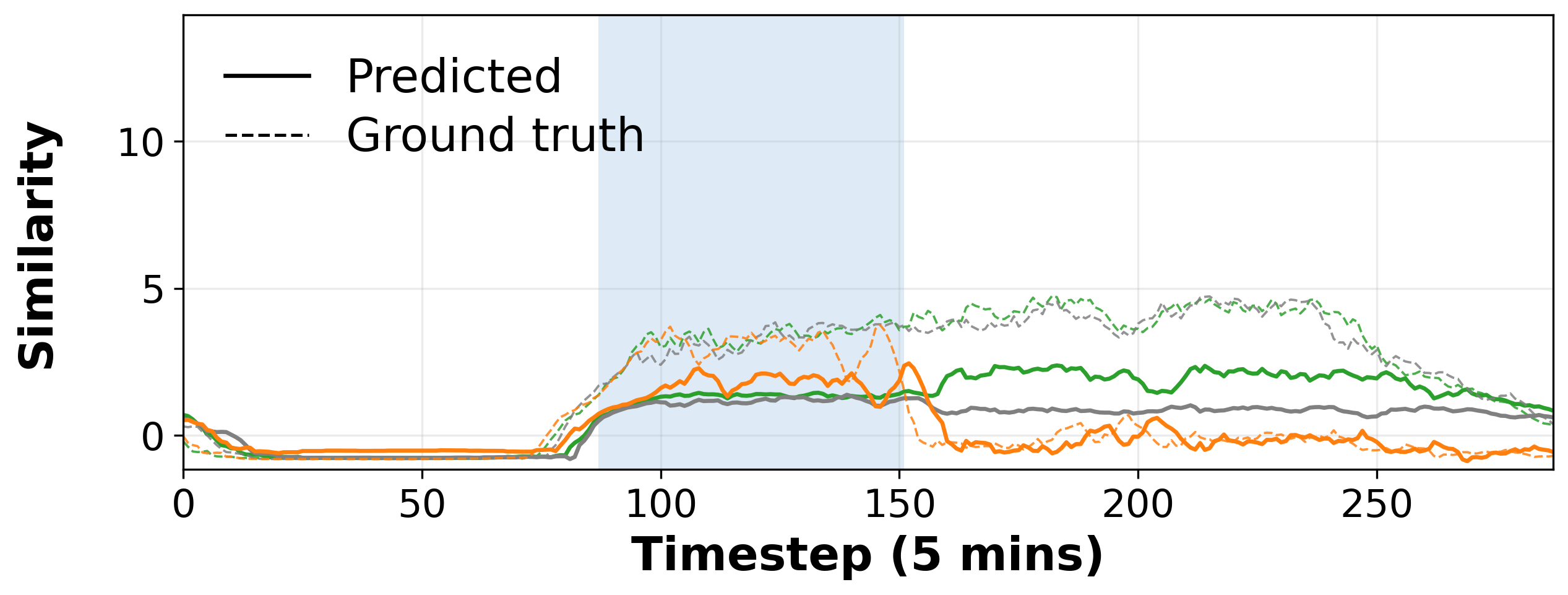}\\[-1mm]

        \includegraphics[width=\linewidth]{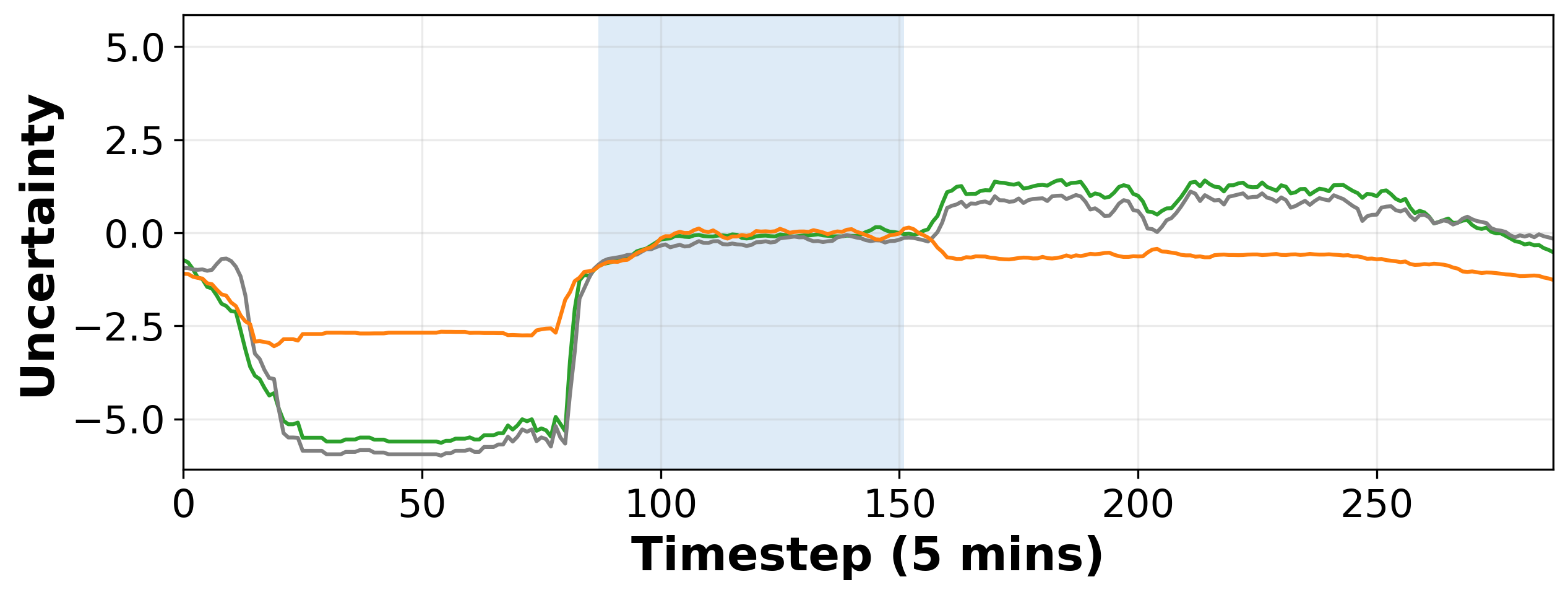}\\[-1mm]

        \includegraphics[width=\linewidth]{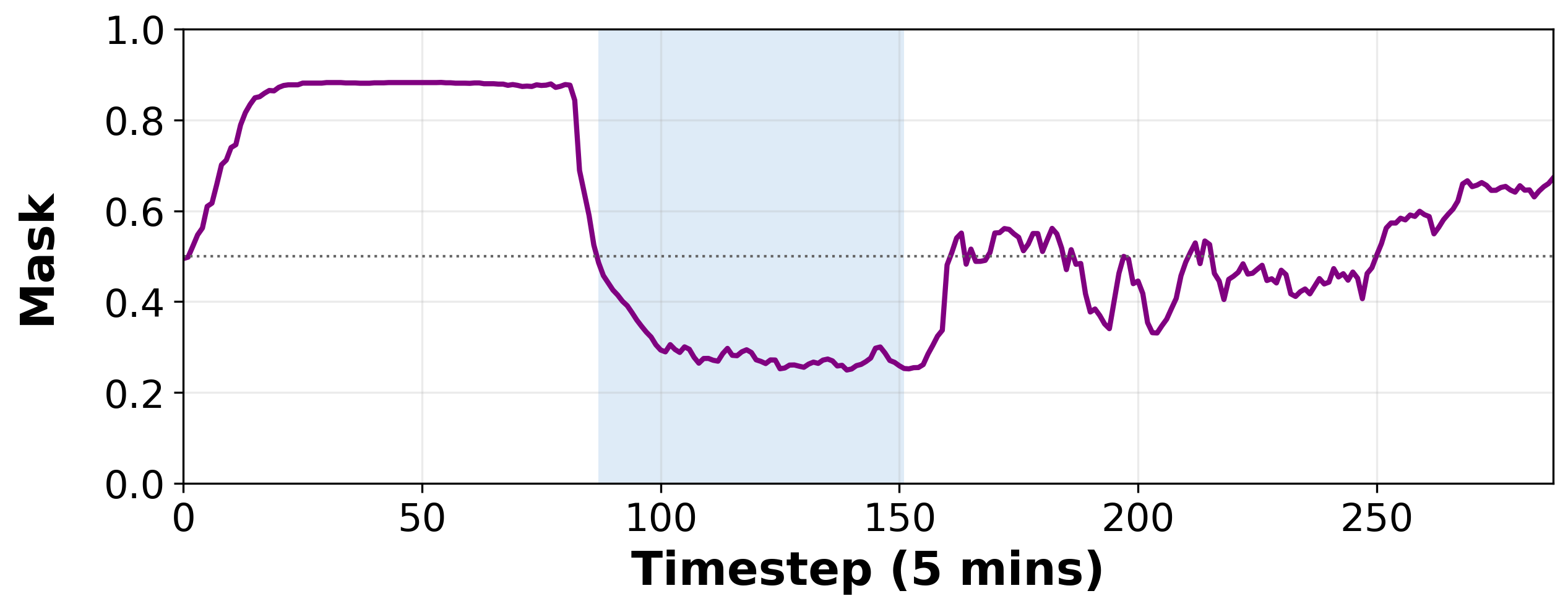}\\[-1mm]

        \includegraphics[width=\linewidth]{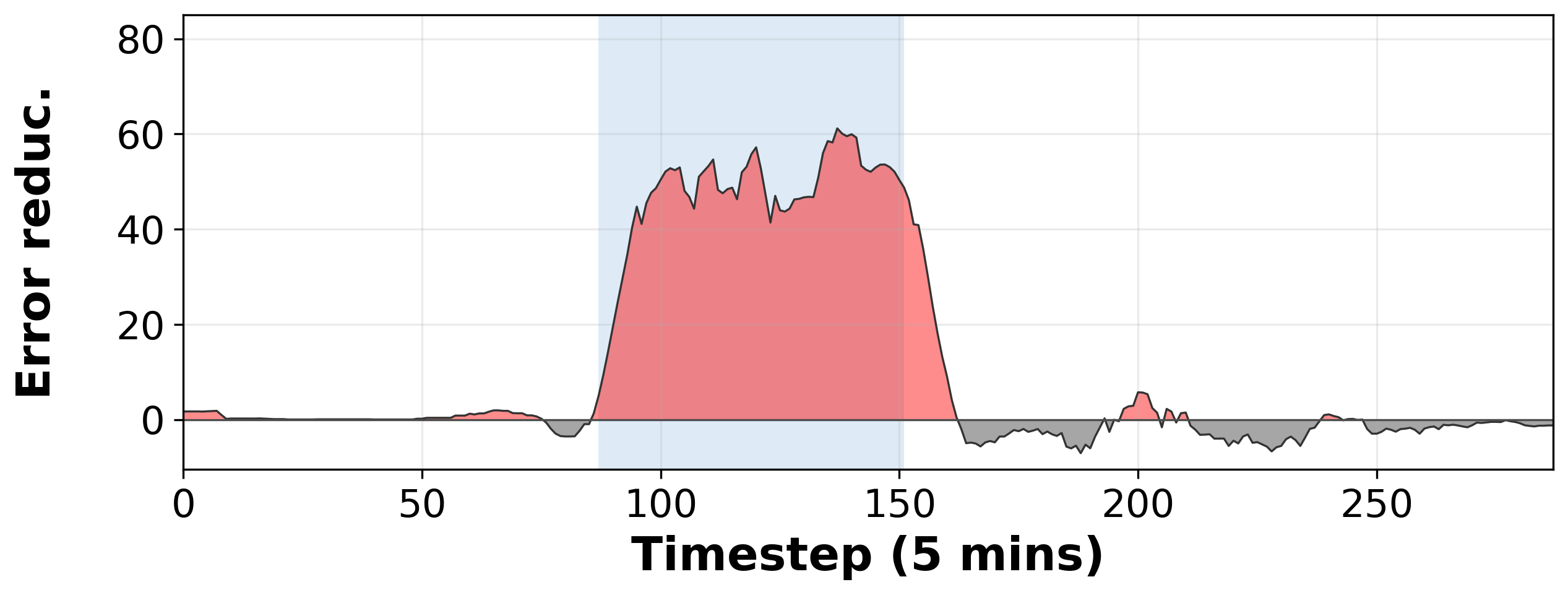}

        \caption{PEMS04 -- Node 247.}
        \label{fig:pems04-node247}
    \end{subfigure}

    \caption{From top to bottom, the panels show the traffic forecasts, predicted similarities $\hat{d}k$ and their ground-truth values $d_k^{+}$, uncertainties $s_k$, value from masking matrix $\mathcal{M}$, and the error reduction over the variant without $\mathcal{M}$. Here, the subscripts $2\mathrm{W}$, $1\mathrm{W}$, and $1\mathrm{D}$ denote the periodic reference patterns from two weeks, one week, and one day earlier, respectively. Blue shading marks intervals where all $d^{+}_{k}$ exceed the 90th percentile and $\mathcal{M}<0.5$.}
    \label{fig_detection}
\end{figure}

In this section, we visualize how \algshort{} detects irregular traffic patterns through its internal variables: the predicted similarity $\hat{d_{k}}$, uncertainty $s_{k}$ and masking matrix $\mathcal{M}$. We also investigate whether the performance improvement is concentrated in such irregular situations. Fig.~\ref{fig_detection} presents two representative cases. In periodic regimes, \algshort{} accurately predicts the similarity between the current pattern and each periodic reference. When traffic becomes aperiodic, the predicted similarities may deviate from their ground-truth values across all periodic references. In these cases, the model assigns higher uncertainty, resulting in low mask values that suppress unreliable periodic information. Accordingly, the largest error reductions over the variant without $\mathcal{M}$ occur in the highlighted aperiodic intervals, demonstrating the effectiveness of uncertainty-aware masking under irregular traffic conditions.

\newpage

\subsection{Analysis on Uncertainty}

\begin{figure*}[!htbp]
\centering
\begin{subfigure}{0.48\textwidth}
\centering
\includegraphics[width=\linewidth]{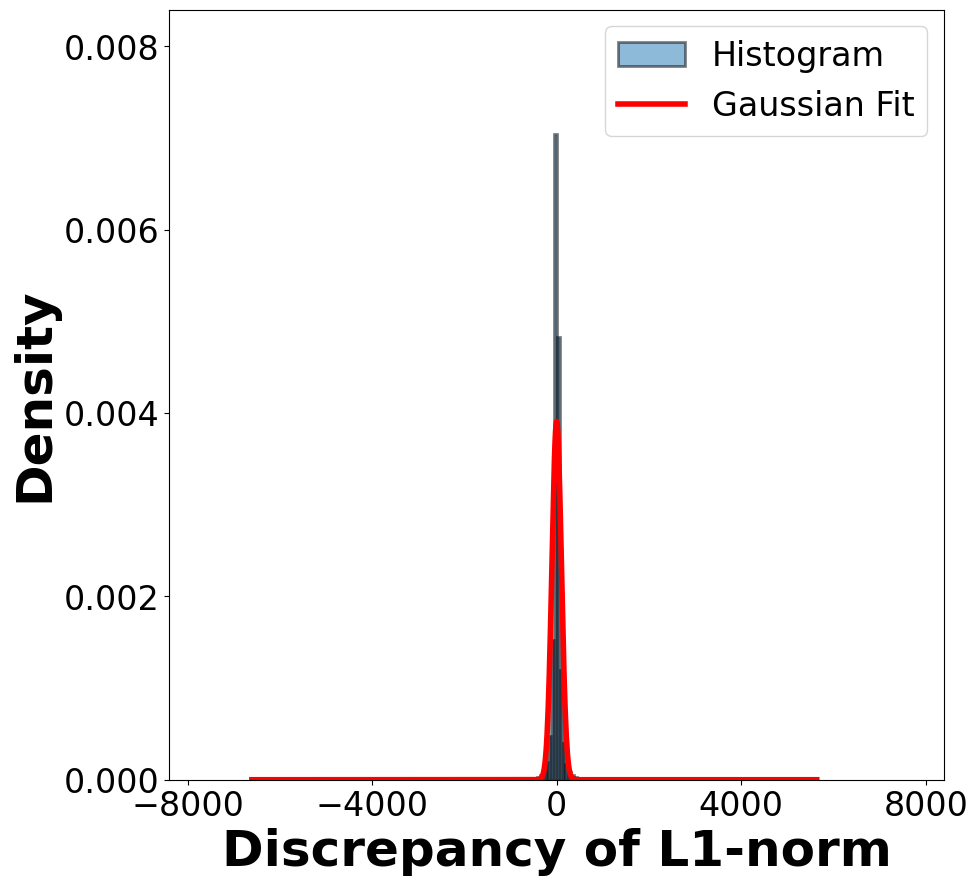}
\caption{PEMS03.}
\end{subfigure}
\begin{subfigure}{0.48\textwidth}
\centering
\includegraphics[width=\linewidth]{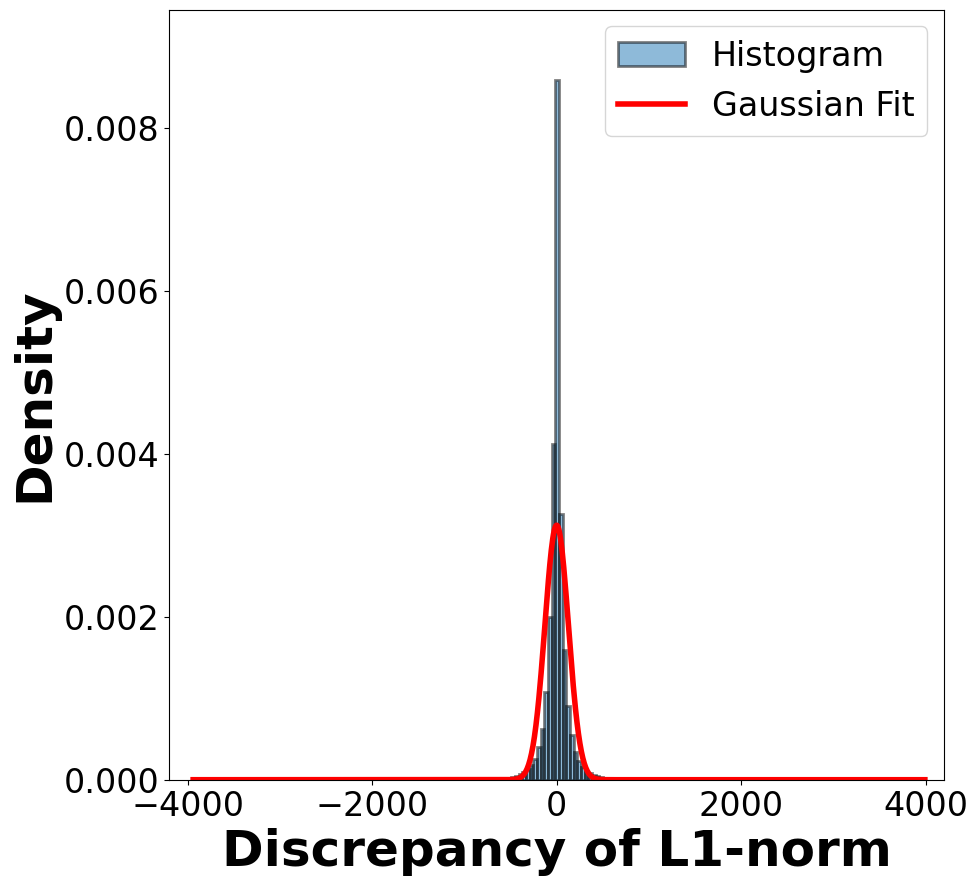}
\caption{PEMS04.}
\end{subfigure}
\begin{subfigure}{0.48\textwidth}
\centering
\includegraphics[width=\linewidth]{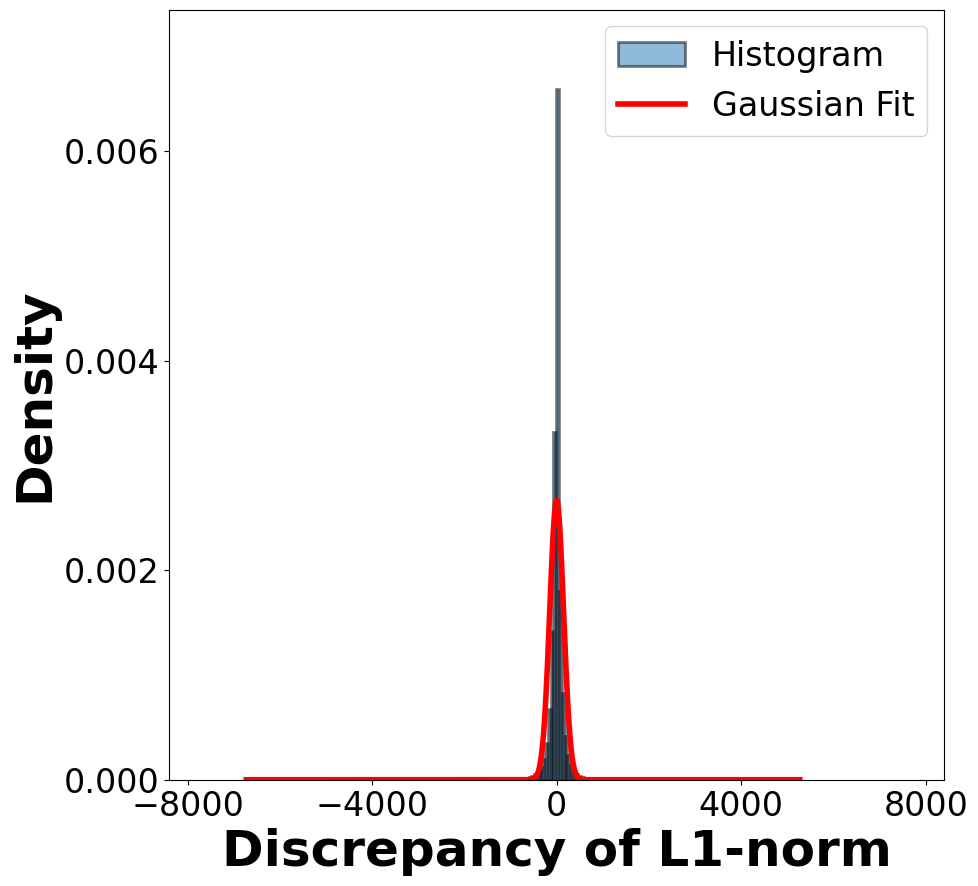}
\caption{PEMS07.}
\end{subfigure}
\begin{subfigure}{0.48\textwidth}
\centering
\includegraphics[width=\linewidth]{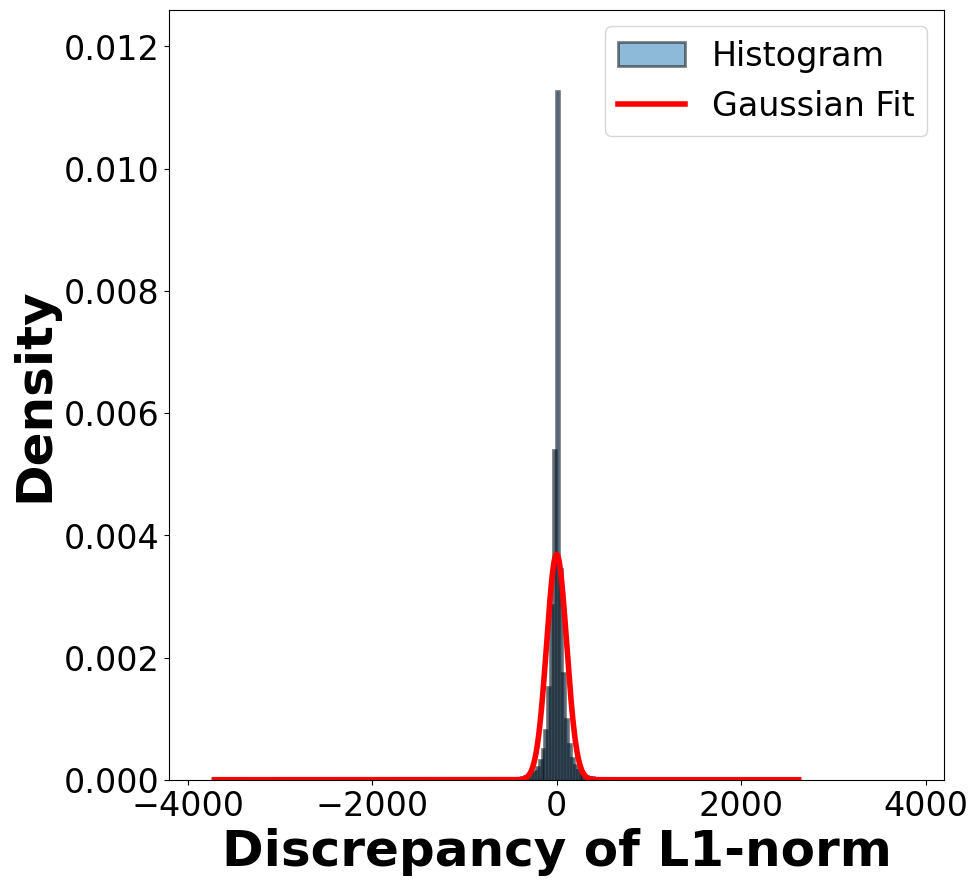}
\caption{PEMS08.}
\end{subfigure}
\caption{L1-norm discrepancy distribution.}
\label{ap:discrepancy}
\end{figure*}

In this section, we clarify and justify the use of uncertainty to model the discrepancy between the L1-norm of current data and historical data and the L1-norm of target data and its corresponding historical data. We adopt \textit{aleatoric uncertainty} from \cite{uncertainty} to model such discrepancy. In \cite{uncertainty}, aleatoric uncertainty is introduced to capture the inherent stochastic noise in observations that cannot be completely eliminated even with sufficient data. Following this perspective, we interpret the discrepancy between the input-historical L1-norm and the target-historical L1-norm as an intrinsic stochastic variability in historical similarity estimation.

Fig.~\ref{ap:discrepancy} illustrates the discrepancy distributions across all datasets. The distributions are consistently centered around zero while exhibiting non-negligible variance, indicating that historical similarity generally provides meaningful information but is not perfectly deterministic. Even when the current input exhibits periodic patterns similar to historical observations, the target may deviate from the corresponding historical behavior due to irregular traffic dynamics and stochastic temporal variations.

Furthermore, the discrepancy distributions exhibit approximately Gaussian-like shapes centered around zero, which is consistent with the probabilistic assumption commonly adopted in aleatoric uncertainty modeling~\cite{uncertainty}. Therefore, the discrepancy can be interpreted as an input-dependent stochastic noise arising from the imperfect correspondence between current-historical similarity and target-historical similarity. Based on this interpretation, the proposed uncertainty module estimates the reliability of historical similarity information, where larger uncertainty indicates that the historical similarity relationship is less reliable for forecasting the future target.

\section{Modularity of SIMBAD}
\label{ap:modularization}
In this section, we show the full results of modularization of \algshort{} discussed in \S 4.5 on all baselines and metrics. Table \ref{modularization_appendix} below shows results of baseline models with and without the proposed Aperiodicity-Aware Modulation Layer. Among the 120 results, the proposed module improves 77 cases, yielding an average relative improvement of 2.00\%. Seven of the ten backbones achieve positive average gains, with substantial improvements in DCRNN~\cite{dcrnn} and STSGCN~\cite{stsgcn} with average error reductions of 7.97\% and 7.87\%, respectively. PEMS07 and PEMS08 gain largest average improvements, with an average of 4.71\% and 3.78\%, respectively. These results demonstrate that the proposed Aperiodicity-Aware Modulation Layer can effectively enhance diverse spatio-temporal forecasting architectures.

\begin{table*}[!htbp]
\centering
\caption{Performance comparison of baseline models with and without
the proposed Aperiodicity-Aware Modulation Layer on PEMS datasets. Superscript on +Ours indicate improvement (\textcolor{red}{$\blacktriangle$}) or degradation (\textcolor{blue}{$\blacktriangledown$}).}
\label{modularization_appendix}
\providecommand{\upmark}{%
  \rlap{\hspace{0.12em}\textsuperscript{\textcolor{red}{$\blacktriangle$}}}%
}
\providecommand{\downmark}{%
  \rlap{\hspace{0.12em}\textsuperscript{\textcolor{blue}{$\blacktriangledown$}}}%
}
\vspace{-2mm}
\resizebox{\textwidth}{!}{
\begin{tabular}{@{}l ccc ccc ccc ccc@{}}
\toprule
\textsc{Model}
& \multicolumn{3}{c}{PEMS03}
& \multicolumn{3}{c}{PEMS04}
& \multicolumn{3}{c}{PEMS07}
& \multicolumn{3}{c}{PEMS08} \\
\cmidrule(lr){2-4}
\cmidrule(lr){5-7}
\cmidrule(lr){8-10}
\cmidrule(lr){11-13}
\textsc{Metric}
& MAE & RMSE & MAPE\%
& MAE & RMSE & MAPE\%
& MAE & RMSE & MAPE\%
& MAE & RMSE & MAPE\% \\
\midrule

GWNET
& 14.94 & 26.06 & 15.52
& 19.23 & 30.76 & 14.04
& 20.63 & 33.26 & 9.14
& 15.06 & 23.79 & 9.73 \\
+ Ours
& 15.96\downmark & 29.15\downmark & 16.55\downmark
& 18.80\upmark & 30.68\upmark & 13.17\upmark
& 19.11\upmark & 31.71\upmark & 8.40\upmark
& 14.10\upmark & 22.91\upmark & 9.61\upmark \\

DCRNN
& 16.67 & 29.06 & 16.60
& 21.68 & 34.19 & 15.05
& 23.13 & 36.13 & 10.18
& 16.81 & 25.91 & 11.71 \\
+ Ours
& 15.91\upmark & 27.80\upmark & 16.51\upmark
& 20.12\upmark & 32.18\upmark & 13.75\upmark
& 19.99\upmark & 32.68\upmark & 8.55\upmark
& 15.62\upmark & 24.70\upmark & 10.11\upmark \\

AGCRN
& 16.21 & 28.57 & 15.35
& 19.85 & 32.96 & 13.38
& 21.35 & 35.12 & 9.02
& 16.54 & 26.23 & 10.47 \\
+ Ours
& 16.48\downmark & 29.08\downmark & 16.70\downmark
& 18.92\upmark & 30.64\upmark & 13.73\downmark
& 19.65\upmark & 32.83\upmark & 8.42\upmark
& 14.33\upmark & 23.29\upmark & 9.81\upmark \\

STSGCN
& 18.24 & 30.51 & 17.37
& 21.46 & 34.07 & 14.37
& 24.07 & 38.69 & 10.17
& 17.05 & 26.47 & 10.87 \\
+ Ours
& 17.40\upmark & 30.71\downmark & 17.97\downmark
& 19.34\upmark & 31.36\upmark & 13.28\upmark
& 20.35\upmark & 33.97\upmark & 8.62\upmark
& 15.26\upmark & 24.49\upmark & 10.04\upmark \\

STAEformer
& 16.67 & 29.56 & 17.21
& 18.63 & 30.86 & 13.10
& 19.91 & 33.94 & 8.58
& 13.82 & 23.55 & 9.23 \\
+ Ours
& 16.03\upmark & 28.30\upmark & 17.28\downmark
& 18.69\downmark & 30.35\upmark & 13.64\downmark
& 19.82\upmark & 33.08\upmark & 8.79\downmark
& 13.64\upmark & 22.99\upmark & 9.58\downmark \\

DSTAGNN
& 15.99 & 28.20 & 15.45
& 19.85 & 31.91 & 13.42
& 21.84 & 34.95 & 9.58
& 16.19 & 25.44 & 10.61 \\
+ Ours
& 16.44\downmark & 28.39\downmark & 18.39\downmark
& 19.53\upmark & 31.31\upmark & 14.58\downmark
& 21.14\upmark & 34.00\upmark & 9.68\downmark
& 14.89\upmark & 23.58\upmark & 10.73\downmark \\

STWave
& 15.91 & 28.06 & 16.62
& 19.63 & 32.02 & 13.66
& 21.32 & 35.68 & 8.99
& 14.01 & 23.94 & 9.33 \\
+ Ours
& 16.00\downmark & 28.74\downmark & 16.81\downmark
& 18.84\upmark & 30.78\upmark & 13.05\upmark
& 19.30\upmark & 32.20\upmark & 8.28\upmark
& 13.71\upmark & 22.94\upmark & 9.15\upmark \\

STIDGCN
& 15.98 & 26.66 & 17.24
& 18.38 & 30.47 & 12.42
& 19.20 & 32.87 & 8.18
& 13.71 & 23.33 & 9.00 \\
+ Ours
& 15.20\upmark & 25.60\upmark & 15.70\upmark
& 18.27\upmark & 30.23\upmark & 13.02\downmark
& 18.83\upmark & 32.20\upmark & 7.98\upmark
& 13.12\upmark & 22.55\upmark & 8.74\upmark \\

DTRformer
& 15.83 & 27.61 & 16.31
& 18.49 & 30.62 & 13.08
& 19.52 & 32.53 & 8.38
& 13.50 & 22.90 & 9.05 \\
+ Ours
& 17.43\downmark & 31.00\downmark & 17.85\downmark
& 19.46\downmark & 31.65\downmark & 13.70\downmark
& 19.92\downmark & 32.91\downmark & 8.72\downmark
& 14.24\downmark & 23.40\downmark & 9.61\downmark \\

LHMformer
& 16.22 & 27.81 & 16.76
& 18.45 & 30.70 & 13.00
& 19.27 & 32.54 & 8.11
& 13.66 & 23.04 & 9.02 \\
+ Ours
& 16.15\upmark & 28.70\downmark & 16.55\upmark
& 18.83\downmark & 30.52\upmark & 13.41\downmark
& 19.76\downmark & 32.94\downmark & 8.92\downmark
& 13.53\upmark & 22.63\upmark & 9.17\downmark \\

\midrule
\textbf{SIMBAD}
& \textbf{14.77} & \textbf{25.87} & \textbf{14.35}
& \textbf{18.03} & \textbf{30.08} & \textbf{12.36}
& \textbf{17.86} & \textbf{30.99} & \textbf{7.44}
& \textbf{12.58} & \textbf{22.13} & \textbf{8.35} \\

\bottomrule
\end{tabular}%
}
\end{table*}

\newpage

\section{Stretched Training Dataset}
\vspace{-2mm}
As in Section 2, conventional models are given previous observation of $T$ timesteps to predict future traffic flow of $T^{\prime}$ timesteps with input $X \in \mathbb{R}^{N \times f \times T}$ by learning a mapping function $F$
\begin{equation}
    [X_{t-T+1}, \cdots, X_{t};\mathcal{G}] \underset{F}\longrightarrow [X_{t+1}, \cdots, X_{t+T^{\prime}}]
\end{equation}
where the previous observation of $T$ timesteps are referred to as \textit{recent} input in our model. However, our model incorporates extra historical data (\textit{weekly} and \textit{daily}) as inputs for prediction. As a result, our model takes in more inputs than existing baselines to predict the same timesteps of traffic flow. This may result in an imbalance of training datasets between our model and existing ones which may lead to unfair comparison. To overcome this, we stretched the length of training datasets for baselines to match the scope of training data. Table \ref{scope_performance} below shows results between our model and baselines with stretched training data.
\vspace{-2mm}

\begin{table*}[!htbp]
\centering
\caption{Performance comparison on the stretched training dataset.}
\label{scope_performance}
\vspace{-2mm}
\resizebox{\textwidth}{!}{
\begin{tabular}{@{}l ccc ccc ccc ccc@{}}
\toprule
\textsc{Model}
& \multicolumn{3}{c}{PEMS03}
& \multicolumn{3}{c}{PEMS04}
& \multicolumn{3}{c}{PEMS07}
& \multicolumn{3}{c}{PEMS08} \\
\cmidrule(lr){2-4}
\cmidrule(lr){5-7}
\cmidrule(lr){8-10}
\cmidrule(lr){11-13}
\textsc{Metric}
& MAE & RMSE & MAPE\%
& MAE & RMSE & MAPE\%
& MAE & RMSE & MAPE\%
& MAE & RMSE & MAPE\% \\
\midrule

GWNET
& \textbf{14.63} & \textbf{25.45} & \underline{14.71}
& 18.88 & 30.38 & 13.45
& 20.52 & 33.10 & 8.84
& 14.72 & 23.53 & 9.56 \\

DCRNN
& 16.31 & 28.71 & 16.07
& 20.91 & 33.21 & 13.96
& 22.66 & 35.68 & 10.11
& 16.13 & 25.32 & 10.42 \\

AGCRN
& 15.93 & 27.82 & 14.88
& 19.47 & 32.16 & 13.05
& 20.91 & 34.60 & 8.86
& 16.04 & 25.46 & 10.29 \\

STSGCN
& 17.65 & 29.39 & 16.73
& 20.97 & 33.61 & 13.88
& 23.78 & 38.72 & 10.01
& 16.64 & 25.98 & 10.68 \\

STAEformer
& 15.81 & 27.93 & 16.12
& 18.41 & 30.87 & 12.83
& 19.73 & 33.53 & 8.72
& 13.70 & 23.19 & 9.33 \\

DSTAGNN
& 15.64 & 27.58 & 14.78
& 19.41 & 31.60 & 12.95
& 21.62 & 34.72 & 9.45
& 15.81 & 24.90 & 10.08 \\

STWave
& 15.03 & 27.14 & 15.21
& 18.75 & 30.83 & 12.97
& 20.53 & 34.33 & 8.76
& 13.45 & 23.39 & 8.90 \\

STIDGCN
& 15.50 & 25.91 & 17.05
& 18.39 & 30.52 & \underline{12.69}
& \underline{19.02} & 32.74 & \underline{7.99}
& 13.50 & 23.21 & 9.00 \\

DTRformer
& 15.27 & 27.14 & 15.64
& 18.21 & \underline{30.11} & 12.75
& 19.32 & \underline{32.31} & 8.25
& 13.35 & 22.80 & 8.90 \\

LHMformer
& 15.49 & 25.88 & 15.82
& \underline{18.17} & 30.62 & 12.75
& 19.14 & 32.44 & 8.04
& \underline{13.32} & \underline{22.78} & \underline{8.78} \\

ASTGCN
& 20.25 & 35.99 & 20.67
& 22.20 & 35.24 & 15.66
& 25.50 & 39.75 & 11.43
& 17.07 & 26.49 & 11.80 \\

\midrule
\textbf{SIMBAD}
& \underline{14.77} & \underline{25.87} & \textbf{14.35}
& \textbf{18.03} & \textbf{30.08} & \textbf{12.36}
& \textbf{17.86} & \textbf{30.99} & \textbf{7.44}
& \textbf{12.58} & \textbf{22.13} & \textbf{8.35} \\

\bottomrule
\end{tabular}
}
\end{table*}

\section{Standard Deviation Results}
In this section, we show standard deviation results of the overall performance results. We also compare with NLinear and DLinear~\cite{linear}, which have been shown to be competitive on traffic data with minimal parameters.
\vspace{-2mm}

\begin{table*}[!htbp]
\centering
\caption{Standard deviation results on PEMS datasets.}
\label{tab:overall_ablation_std}
\vspace{-2mm}
\resizebox{\textwidth}{!}{
\begin{tabular}{@{}l ccc ccc ccc ccc@{}}
\toprule
\textsc{Model}
& \multicolumn{3}{c}{PEMS03}
& \multicolumn{3}{c}{PEMS04}
& \multicolumn{3}{c}{PEMS07}
& \multicolumn{3}{c}{PEMS08} \\
\cmidrule(lr){2-4}
\cmidrule(lr){5-7}
\cmidrule(lr){8-10}
\cmidrule(lr){11-13}
\textsc{Metric}
& MAE & RMSE & MAPE\%
& MAE & RMSE & MAPE\%
& MAE & RMSE & MAPE\%
& MAE & RMSE & MAPE\% \\
\midrule

NLinear
& \makecell{21.07 \\ $\pm$ 0.00} & \makecell{34.07 \\ $\pm$ 0.01} & \makecell{19.50 \\ $\pm$ 0.02}
& \makecell{28.80 \\ $\pm$ 0.00} & \makecell{45.47 \\ $\pm$ 0.01} & \makecell{19.00 \\ $\pm$ 0.01}
& \makecell{31.28 \\ $\pm$ 0.00} & \makecell{49.12 \\ $\pm$ 0.01} & \makecell{13.17 \\ $\pm$ 0.01}
& \makecell{22.76 \\ $\pm$ 0.00} & \makecell{36.12 \\ $\pm$ 0.01} & \makecell{13.57 \\ $\pm$ 0.01} \\

DLinear
& \makecell{20.72 \\ $\pm$ 0.01} & \makecell{33.28 \\ $\pm$ 0.02} & \makecell{20.73 \\ $\pm$ 0.14}
& \makecell{28.34 \\ $\pm$ 0.00} & \makecell{44.44 \\ $\pm$ 0.01} & \makecell{20.63 \\ $\pm$ 0.07}
& \makecell{30.66 \\ $\pm$ 0.00} & \makecell{47.77 \\ $\pm$ 0.03} & \makecell{13.31 \\ $\pm$ 0.05}
& \makecell{22.44 \\ $\pm$ 0.01} & \makecell{35.31 \\ $\pm$ 0.02} & \makecell{13.78 \\ $\pm$ 0.11} \\

GWNET
& \makecell{\underline{14.94} \\ $\pm$ 0.12} & \makecell{\underline{26.06} \\ $\pm$ 0.38} & \makecell{15.52 \\ $\pm$ 0.79}
& \makecell{19.23 \\ $\pm$ 0.09} & \makecell{30.76 \\ $\pm$ 0.13} & \makecell{14.04 \\ $\pm$ 0.51}
& \makecell{20.63 \\ $\pm$ 0.05} & \makecell{33.26 \\ $\pm$ 0.07} & \makecell{9.14 \\ $\pm$ 0.17}
& \makecell{15.06 \\ $\pm$ 0.11} & \makecell{23.79 \\ $\pm$ 0.12} & \makecell{9.73 \\ $\pm$ 0.02} \\

DCRNN
& \makecell{16.67 \\ $\pm$ 0.12} & \makecell{29.06 \\ $\pm$ 0.33} & \makecell{16.60 \\ $\pm$ 0.93}
& \makecell{21.68 \\ $\pm$ 0.18} & \makecell{34.19 \\ $\pm$ 0.22} & \makecell{15.05 \\ $\pm$ 0.69}
& \makecell{23.13 \\ $\pm$ 0.13} & \makecell{36.13 \\ $\pm$ 0.15} & \makecell{10.18 \\ $\pm$ 0.14}
& \makecell{16.81 \\ $\pm$ 0.07} & \makecell{25.91 \\ $\pm$ 0.08} & \makecell{11.71 \\ $\pm$ 0.58} \\

AGCRN
& \makecell{16.21 \\ $\pm$ 0.12} & \makecell{28.57 \\ $\pm$ 0.39} & \makecell{\underline{15.35} \\ $\pm$ 0.38}
& \makecell{19.85 \\ $\pm$ 0.14} & \makecell{32.96 \\ $\pm$ 0.41} & \makecell{13.38 \\ $\pm$ 0.24}
& \makecell{21.35 \\ $\pm$ 0.10} & \makecell{35.12 \\ $\pm$ 0.13} & \makecell{9.02 \\ $\pm$ 0.07}
& \makecell{16.54 \\ $\pm$ 0.25} & \makecell{26.23 \\ $\pm$ 0.36} & \makecell{10.47 \\ $\pm$ 0.19} \\

STSGCN
& \makecell{18.24 \\ $\pm$ 0.11} & \makecell{30.51 \\ $\pm$ 0.52} & \makecell{17.37 \\ $\pm$ 0.17}
& \makecell{21.46 \\ $\pm$ 0.06} & \makecell{34.07 \\ $\pm$ 0.14} & \makecell{14.37 \\ $\pm$ 0.07}
& \makecell{24.07 \\ $\pm$ 0.14} & \makecell{38.69 \\ $\pm$ 0.34} & \makecell{10.17 \\ $\pm$ 0.07}
& \makecell{17.05 \\ $\pm$ 0.17} & \makecell{26.47 \\ $\pm$ 0.25} & \makecell{10.87 \\ $\pm$ 0.11} \\

STAEformer
& \makecell{16.67 \\ $\pm$ 0.22} & \makecell{29.56 \\ $\pm$ 0.42} & \makecell{17.21 \\ $\pm$ 0.33}
& \makecell{18.63 \\ $\pm$ 0.03} & \makecell{30.86 \\ $\pm$ 0.25} & \makecell{13.10 \\ $\pm$ 0.17}
& \makecell{19.91 \\ $\pm$ 0.17} & \makecell{33.94 \\ $\pm$ 0.49} & \makecell{8.58 \\  $\pm$ 0.08}
& \makecell{13.82 \\ $\pm$ 0.09} & \makecell{23.55 \\ $\pm$ 0.23} & \makecell{9.23 \\ $\pm$ 0.14} \\

DSTAGNN
& \makecell{15.99 \\ $\pm$ 0.11} & \makecell{28.20 \\ $\pm$ 0.15} & \makecell{15.45 \\ $\pm$ 0.71}
& \makecell{19.85 \\ $\pm$ 0.19} & \makecell{31.91 \\ $\pm$ 0.11} & \makecell{13.42 \\ $\pm$ 0.30}
& \makecell{21.84 \\ $\pm$ 0.30} & \makecell{34.95 \\ $\pm$ 0.36} & \makecell{9.58 \\ $\pm$ 0.32}
& \makecell{16.19 \\ $\pm$ 0.18} & \makecell{25.44 \\ $\pm$ 0.25} & \makecell{10.61 \\ $\pm$ 0.27} \\

STWave
& \makecell{15.91 \\ $\pm$ 0.34} & \makecell{28.06 \\ $\pm$ 0.16} & \makecell{16.62 \\ $\pm$ 0.34}
& \makecell{19.63 \\ $\pm$ 0.51} & \makecell{32.02 \\ $\pm$ 0.74} & \makecell{13.66 \\ $\pm$ 0.38}
& \makecell{21.32 \\ $\pm$ 0.37} & \makecell{35.68 \\ $\pm$ 0.75} & \makecell{8.99 \\ $\pm$ 0.13}
& \makecell{14.01 \\ $\pm$ 0.39} & \makecell{23.94 \\ $\pm$ 0.71} & \makecell{9.33 \\ $\pm$ 0.42} \\

STIDGCN
& \makecell{15.98 \\ $\pm$ 0.11} & \makecell{26.66 \\ $\pm$ 0.18} & \makecell{17.24 \\ $\pm$ 0.24}
& \makecell{\underline{18.38} \\ $\pm$ 0.06} & \makecell{\underline{30.47} \\ $\pm$ 0.18} & \makecell{\underline{12.42} \\ $\pm$ 0.21}
& \makecell{\underline{19.20} \\ $\pm$ 0.06} & \makecell{32.87 \\ $\pm$ 0.11} & \makecell{8.18 \\ $\pm$ 0.23}
& \makecell{13.71 \\ $\pm$ 0.10} & \makecell{23.33 \\ $\pm$ 0.03} & \makecell{\underline{9.00} \\ $\pm$ 0.08} \\

DTRformer
& \makecell{15.83 \\ $\pm$ 0.16} & \makecell{27.61 \\ $\pm$ 0.42} & \makecell{16.31 \\ $\pm$ 0.28}
& \makecell{18.49 \\ $\pm$ 0.10} & \makecell{30.62 \\ $\pm$ 0.29} & \makecell{13.08 \\ $\pm$ 0.18}
& \makecell{19.52 \\ $\pm$ 0.11} & \makecell{\underline{32.53} \\ $\pm$ 0.14} & \makecell{8.38 \\ $\pm$ 0.14}
& \makecell{\underline{13.50} \\ $\pm$ 0.04} & \makecell{\underline{22.90} \\ $\pm$ 0.09} & \makecell{9.05 \\ $\pm$ 0.17} \\

LHMformer
& \makecell{16.22 \\ $\pm$ 0.09} & \makecell{27.81 \\ $\pm$ 0.25} & \makecell{16.76 \\ $\pm$ 0.22}
& \makecell{18.45 \\ $\pm$ 0.03} & \makecell{30.70 \\ $\pm$ 0.08} & \makecell{13.00 \\ $\pm$ 0.05}
& \makecell{19.27 \\ $\pm$ 0.03} & \makecell{32.54 \\ $\pm$ 0.05} & \makecell{\underline{8.11} \\ $\pm$ 0.01}
& \makecell{13.66 \\ $\pm$ 0.07} & \makecell{23.04 \\ $\pm$ 0.02} & \makecell{9.02 \\ $\pm$ 0.06} \\

ASTGCN
& \makecell{20.25 \\ $\pm$ 0.27} & \makecell{35.99 \\ $\pm$ 0.53} & \makecell{20.67 \\ $\pm$ 0.55}
& \makecell{22.20 \\ $\pm$ 0.26} & \makecell{35.24 \\ $\pm$ 0.49} & \makecell{15.66 \\ $\pm$ 0.59}
& \makecell{25.50 \\ $\pm$ 0.37} & \makecell{39.75 \\ $\pm$ 0.66} & \makecell{11.43 \\ $\pm$ 0.15}
& \makecell{17.07 \\ $\pm$ 0.21} & \makecell{26.49 \\ $\pm$ 0.32} & \makecell{11.80 \\ $\pm$ 0.34} \\

\midrule
\textbf{SIMBAD}
& \makecell{\textbf{14.77} \\ $\pm$ 0.04} & \makecell{\textbf{25.87} \\ $\pm$ 0.15} & \makecell{\textbf{14.35} \\ $\pm$ 0.13}
& \makecell{\textbf{18.03} \\ $\pm$ 0.04} & \makecell{\textbf{30.08} \\ $\pm$ 0.06} & \makecell{\textbf{12.36} \\ $\pm$ 0.02}
& \makecell{\textbf{17.86} \\ $\pm$ 0.02} & \makecell{\textbf{30.99} \\ $\pm$ 0.02} & \makecell{\textbf{7.44} \\ $\pm$ 0.02}
& \makecell{\textbf{12.58} \\ $\pm$ 0.02} & \makecell{\textbf{22.13} \\ $\pm$ 0.03} & \makecell{\textbf{8.35} \\ $\pm$ 0.07} \\

\bottomrule
\end{tabular}
}
\end{table*}

\section{Hyperparameter Analysis}
\label{ap:hyperparameter}
In this section, we compare two hyperparameter settings. (1) $\tau$: a temperature parameter which controls the sharpness of distribution of Eq.~\ref{distribution}. (2) $\lambda_{u}$: regularizes the loss of Eq.~\ref{unc_loss}. Table \ref{hyperparameter} shows results in terms of MAE, RMSE, and MAPE on all four datasets. Our choice of $\tau$ is 0.2 and $\lambda_u$ is 0.1.

\begin{table}[h]
    \small
    \centering
    \caption{Performance of hyperparameter-tuned variants. Best results are marked in bold.}
    \vspace{1.5mm}
    \label{hyperparameter}
    \begin{tabularx}{\columnwidth}{@{\extracolsep{\fill}} l|YYY|YYY}
        \toprule
        \multirow{2}{*}{}
        & \multicolumn{3}{c|}{\textbf{PEMS03}}
        & \multicolumn{3}{c}{\textbf{PEMS04}} \\
        \cmidrule(lr){2-4} \cmidrule(lr){5-7}
        & MAE & RMSE & MAPE
        & MAE & RMSE & MAPE \\
        \midrule
            Default & \textbf{14.77} & \textbf{25.87} & \textbf{14.35} & 18.03 & 30.08 & 12.36 \\
        \hline
        \textit{\textcolor{gray}{temperature ($\tau$)}} \\
        0.1 & 14.99 & 26.27 & 14.84 & 18.04 & 30.11 & 12.38 \\
        0.3 & 14.99 & 26.30 & 14.66 & 18.03 & 30.08 & 12.37 \\
        0.4 & 15.03 & 26.35 & 14.96 & 18.04 & 30.08 & 12.35 \\
        \hline
        \textit{\textcolor{gray}{regularizer ($\lambda_u$)}} \\
        0.2 & 14.92 & 26.30 & 14.62 & 18.04 & 30.09 & 12.40 \\
        0.3 & 14.98 & 26.38 & 14.79 & \textbf{18.02} & \textbf{30.06} & \textbf{12.32} \\
        0.4 & 14.90 & 26.28 & 14.61 & 18.06 & 30.09 & 12.36 \\
        \bottomrule
    \end{tabularx}
    
    \begin{tabularx}{\columnwidth}{@{\extracolsep{\fill}} l|YYY|YYY}
    \toprule
    \multirow{2}{*}{}
    & \multicolumn{3}{c|}{\textbf{PEMS07}}
    & \multicolumn{3}{c}{\textbf{PEMS08}} \\
    \cmidrule(lr){2-4} \cmidrule(lr){5-7}
    & MAE & RMSE & MAPE
    & MAE & RMSE & MAPE \\
    \midrule
    Default & 17.86 & 30.99 & 7.44 & \textbf{12.58} & 22.13 & \textbf{8.35} \\
    \hline
        \textit{\textcolor{gray}{temperature ($\tau$)}} \\
        0.1 & 17.86 & 30.99 & 7.44 & 12.60 & 22.15 & 8.38 \\
        0.3 & \textbf{17.85} & \textbf{30.98} & \textbf{7.43} & 12.60 & \textbf{22.10} & 8.40 \\
        0.4 & 17.86 & \textbf{30.98} & 7.44 & 12.62 & 22.12 & 8.43 \\
        \hline
        \textit{\textcolor{gray}{regularizer ($\lambda_u$)}} \\
        0.2 & 17.86 & 30.99 & 7.45 & 12.61 & 22.15 & 8.43 \\
        0.3 & 17.88 & 31.00 & 7.47 & 12.63 & 22.15 & 8.46 \\
        0.4 & 17.90 & 31.02 & 7.48 & 12.67 & 22.15 & 8.45 \\
    \bottomrule
    \end{tabularx}
\end{table}

\section{Computational Efficiency}
In this section, we compare the computational complexity between baselines on the tested datasets. Compared to most recent baselines~\cite{DTRformer,lhmformer}, \algshort{} achieves more efficient training and inference while maintaining superior performance. This demonstrates that utilization of periodic patterns and uncertainty can improve forecasting accuracy without introducing excessive computational overhead.

\begin{table*}[!htbp]
\centering
\caption{Train per epoch and testing time comparison.}
\label{scope_runtime}
\vspace{2mm}
\resizebox{\textwidth}{!}{
\begin{tabular}{@{}l cc cc cc cc@{}}
\toprule
\textsc{Model}
& \multicolumn{2}{c}{PEMS03}
& \multicolumn{2}{c}{PEMS04}
& \multicolumn{2}{c}{PEMS07}
& \multicolumn{2}{c}{PEMS08} \\
\cmidrule(lr){2-3}
\cmidrule(lr){4-5}
\cmidrule(lr){6-7}
\cmidrule(lr){8-9}
\textsc{Metric}
& Train (s) & Test (s)
& Train (s) & Test (s)
& Train (s) & Test (s)
& Train (s) & Test (s) \\
\midrule

GWNET
& 8.33 & 1.43
& 4.17 & 0.91
& 27.12 & 4.00
& 3.03 & 0.70 \\

DCRNN
& 51.69 & 10.80
& 28.03 & 5.52
& 105.07 & 44.47
& 25.76 & 4.77 \\

AGCRN
& 17.82 & 2.93
& 10.30 & 1.85
& 27.45 & 3.96
& 11.10 & 1.91 \\

STSGCN
& 23.68 & 6.56
& 11.58 & 2.87
& 97.44 & 21.29
& 12.37 & 1.56 \\

STAEformer
& 40.46 & 4.33
& 19.78 & 2.29
& 137.60 & 14.28
& 11.26 & 1.58 \\

DSTAGNN
& 65.79 & 7.99
& 30.09 & 3.69
& 280.68 & 30.01
& 14.42 & 1.91 \\

STWave
& 47.70 & 7.38
& 24.05 & 4.15
& 141.10 & 21.13
& 14.80 & 2.72 \\

STIDGCN
& 23.65 & 3.15
& 25.74 & 1.94
& 187.64 & 19.98
& 9.08 & 1.19 \\

DTRformer
& 61.97 & 6.28
& 30.54 & 3.11
& 267.35 & 27.41
& 18.14 & 2.14 \\

LHMformer
& 421.42 & 8.08
& 45.24 & 3.28
& 228.11 & 27.54
& 18.98 & 2.25 \\

ASTGCN
& 31.11 & 3.01
& 18.51 & 1.55
& 68.63 & 10.39
& 19.16 & 1.18 \\

\midrule
\textbf{SIMBAD}
& 36.36 & 4.97
& 17.15 & 2.58
& 140.84 & 19.00
& 9.55 & 1.74 \\

\bottomrule
\end{tabular}
}
\end{table*}

\newpage

\section{Additional Datasets}
\label{ap:additional_datasets}
We compare the performance of our model with the baselines on additional datasets METR-LA, PEMS-BAY, and a subset of LargeST~\cite{largest} datasets. For the less efficient baselines, we use one year of traffic data from 2019. The details of the datasets are given in Table \ref{dataset_app} and the results are given in Table~\ref{overall_performance_app}. We split the METR-LA and PEMS-BAY datasets into train, validation and test sets in a $7:1:2$ ratio and SD dataset into $6:2:2$ ratio.

\begin{wraptable}{r}{0.50\linewidth}
\vspace{-10pt}
\small
\centering
\caption{Summary of datasets.}
\label{dataset_app}
\begin{tabular}{lccc}
\toprule
\textsc{Dataset} & \textsc{Nodes} & \textsc{Edges} & \textsc{Timesteps}\\
\midrule
METR-LA & 207 & 1,515 & 34,272 \\
PEMS-BAY & 325 & 2,369 & 52,116 \\
SD (2019) & 716 & 17,319 & 35,040 \\
\bottomrule
\end{tabular}
\vspace{-10pt}
\end{wraptable}

\begin{figure}[H]
\centering
\begin{subfigure}{0.49\textwidth}
\centering
\includegraphics[width=\linewidth]{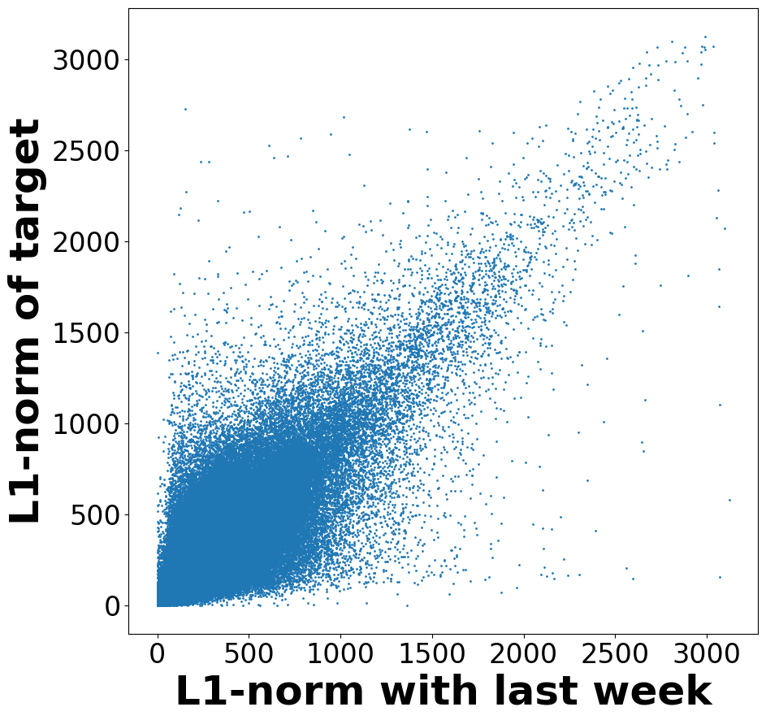}
\caption{PEMS04.}
\end{subfigure}
\hfill
\begin{subfigure}{0.49\textwidth}
\centering
\includegraphics[width=\linewidth]{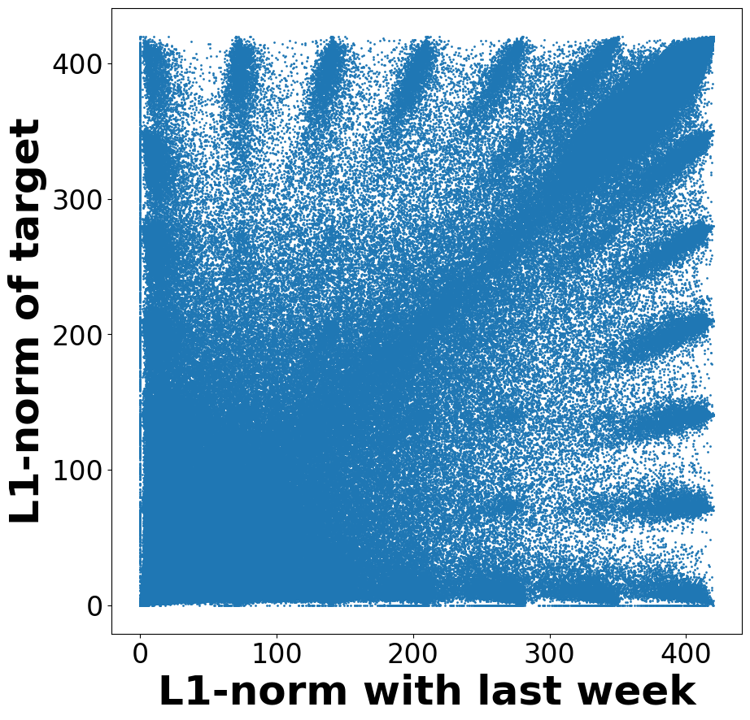}
\caption{METR-LA.}
\end{subfigure}
\caption{Correlation between L1-norms of input and target sequences with respect to their last week counterparts.}
\label{fig_metr}
\end{figure}

\algshort{} achieves improvement of $3.18\%$ on average over the best baselines on the SD dataset and $2.91\%$ on the PEMS-BAY dataset. However, compared to the superior performance on PEMS datasets, \algshort{} shows relatively low performance on METR-LA. In contrast to the clear positive correlation between L1-norm of current-historical and L1-norm of target-historical data of PEMS04 as illustrated in Fig.~\ref{fig_metr}(a), METR-LA shows a substantially weaker correlation between the two L1-norms, as in Fig.~\ref{fig_metr}(b). This suggests that the current state is less informative for selecting reliable historical patterns, violating a key assumption of similarity-based aggregation. As a result, the model's ability to identify and exploit useful periodic information is diminished, leading to reduced performance gains.

\begin{table}[!htbp]
\vspace{1.5mm}
\centering
\caption{Performance comparison on SD (2019), PEMS-BAY, and METR-LA.}
\vspace{1.5mm}
\label{overall_performance_app}
\resizebox{\textwidth}{!}{%
\begin{tabular}{c c c *{12}{c}}
\toprule
\textsc{Dataset} & & \textsc{Metric} &
GWNET & DCRNN & AGCRN & STSGCN & STAEformer & DSTAGNN & STWave & STIDGCN & DTRformer & LHMformer & ASTGCN & \textbf{SIMBAD} \\
\midrule
\multirow{3}{*}{\makecell[c]{SD\\(2019)}} &
\multirow{3}{*}{Average} &
MAE & 19.90 & 21.74 & 23.04 & 204.60 & 18.86 & 21.74 & 19.75 & \underline{18.47} & 19.71 & 19.59 & 23.60 & \textbf{17.98} \\
& & RMSE & 32.63 & 35.63 & 40.69 & 86.63 & 32.99 & 38.28 & 32.35 & \underline{31.30} & 34.30 & 33.33 & 41.48 & \textbf{30.15} \\
& & MAPE & 13.54 & 14.61 & 15.17 & 268.05 & 12.48 & 13.56 & 13.25 & \underline{12.18} & 12.93 & 12.81 & 16.08 & \textbf{11.79} \\
\midrule
\multirow{9}{*}{\rotatebox{90}{PEMS-BAY}} &
\multirow{3}{*}{\makecell[c]{Horizon 3\\(15 min)}} &
MAE & \underline{1.31} & 1.35 & 1.40 & 1.51 & 1.36 & 1.42 & 1.37 & 1.34 & 1.38 & 1.34 & 1.82 & \textbf{1.28} \\
& & RMSE & \underline{2.77} & 2.86 & 2.98 & 3.27 & 2.90 & 3.05 & 2.98 & 2.88 & 2.88 & 2.83 & 3.55 & \textbf{2.76} \\
& & MAPE & \underline{2.74} & 2.85 & 3.00 & 3.77 & 2.90 & 3.12 & 2.93 & 2.86 & 2.96 & 2.85 & 4.09 & \textbf{2.72} \\
\cmidrule{2-15}
& \multirow{3}{*}{\makecell[c]{Horizon 6\\(30 min)}} &
MAE & \underline{1.66} & 1.70 & 1.75 & 1.87 & 1.67 & 1.77 & 1.71 & 1.67 & 1.70 & \underline{1.66} & 2.10 & \textbf{1.58} \\
& & RMSE & \underline{3.78} & 3.90 & 3.97 & 4.31 & 3.86 & 4.05 & 3.97 & 3.87 & 3.78 & 3.79 & 4.26 & \textbf{3.67} \\
& & MAPE & \underline{3.74} & 3.85 & 4.05 & 4.32 & 3.82 & 4.08 & 3.90 & 3.81 & 3.93 & 3.79 & 4.92 & \textbf{3.59} \\
\cmidrule{2-15}
& \multirow{3}{*}{\makecell[c]{Horizon 12\\(60 min)}} &
MAE & 2.00 & 2.05 & 2.11 & 2.30 & 1.96 & 2.12 & 2.02 & 1.94 & 1.99 & \underline{1.93} & 2.43 & \textbf{1.84} \\
& & RMSE & 4.62 & 4.77 & 4.81 & 5.50 & 4.56 & 4.89 & 4.72 & 4.55 & 4.46 & \underline{4.44} & 4.90 & \textbf{4.32} \\
& & MAPE & 4.73 & 4.88 & 5.06 & 5.85 & 4.69 & 5.04 & 4.84 & 4.60 & 4.80 & \underline{4.57} & 5.76 & \textbf{4.40} \\
\midrule
\multirow{9}{*}{\rotatebox{90}{METR-LA}} &
\multirow{3}{*}{\makecell[c]{Horizon 3\\(15 min)}} &
MAE & \underline{2.71} & 2.79 & 3.53 & 3.34 & 2.76 & 3.80 & 2.84 & 2.79 & 2.80 & \textbf{2.66} & 4.94 & 2.79 \\
& & RMSE & \underline{5.18} & 5.37 & 9.25 & 8.22 & 5.41 & 9.38 & 5.65 & 5.54 & 5.49 & \textbf{5.16} & 9.80 & 5.49 \\
& & MAPE & \underline{6.95} & 7.25 & 8.35 & 7.58 & 7.23 & 8.15 & 7.56 & 7.52 & 7.45 & \textbf{6.86} & 9.87 & 7.33 \\
\cmidrule{2-15}
& \multirow{3}{*}{\makecell[c]{Horizon 6\\(30 min)}} &
MAE & 3.10 & 3.20 & 4.56 & 4.12 & \underline{3.09} & 4.81 & 3.24 & 3.16 & 3.14 & \textbf{2.97} & 6.39 & 3.18 \\
& & RMSE & \underline{6.23} & 6.50 & 11.70 & 10.39 & 6.39 & 12.22 & 6.76 & 6.60 & 6.46 & \textbf{6.06} & 12.04 & 6.55 \\
& & MAPE & \underline{8.43} & 8.75 & 10.60 & 9.63 & 8.57 & 10.17 & 9.25 & 9.05 & 8.86 & \textbf{8.11} & 12.05 & 8.88 \\
\cmidrule{2-15}
& \multirow{3}{*}{\makecell[c]{Horizon 12\\(60 min)}} &
MAE & 3.57 & 3.73 & 6.04 & 5.16 & \underline{3.48} & 6.29 & 3.61 & 3.52 & 3.51 & \textbf{3.34} & 6.39 & 3.66 \\
& & RMSE & \underline{7.39} & 7.78 & 14.70 & 13.38 & \textbf{7.05} & 15.23 & 7.70 & 7.51 & \underline{7.39} & \textbf{7.05} & 14.76 & 7.57 \\
& & MAPE & \underline{10.10} & 10.73 & 13.50 & 11.90 & 10.14 & 12.73 & 10.86 & 10.54 & 10.34 & \textbf{9.63} & 14.44 & 10.66 \\
\bottomrule
\end{tabular}%
}
\end{table}

\section{Baseline Details}
\label{ap:baseline}

\begin{itemize}[leftmargin=*]
    \item \textbf{GWNET}~\cite{gwnet} combines gated temporal convolution layer (TCN) and adaptive graph, which is learned from node embeddings, based on GCN. Long-range temporal sequences are handled by stacked dilated 1D causal convolutions.
    \item \textbf{DCRNN}~\cite{dcrnn} leverages diffusion process on a graph and capture spatial dependencies via bidirectional random walks. Temporal dependencies are handled by Gated Recurrent Units (GRUs), which are replaced by the diffusion convolutions to effectively capture spatio-temporal dependencies.
    \item \textbf{AGCRN}~\cite{agcrn} integrates graph convolution with recurrent modeling to capture spatio-temporal dependencies in traffic data. 
    \item \textbf{STSGCN}~\cite{stsgcn} synchronously captures localized spatio-temporal correlations by constructing a single localized spatio-temporal graph that links nodes across adjacent timesteps.
    \item \textbf{STAEformer}~\cite{stae} designed novel spatio-temporal adaptive embeddings which are fed into stacked temporal and spatial vanilla Transformer layers.
    \item \textbf{DSTAGNN}~\cite{dstagnn} utilizes a data-driven graph constructed via Wasserstein distance and employs graph convolutions and multi-scale gated convolutions (M-GTU) to capture diverse spatial and temporal dynamics, respectively.
    \item \textbf{STWave}~\cite{stwave} disentangled trend and event components with wavelet transform to capture both high- and low-frequency components
    \item \textbf{STIDGCN}~\cite{stidgcn} incorporates a traffic pattern bank due to spatial heterogeneity and generates a data-driven dynamic graph.
    \item \textbf{DTRformer}~\cite{DTRformer} utilizes adaptive embedding and cross spatial-temporal attention to fuse dynamic and static information.
    \item \textbf{LHMformer}~\cite{lhmformer} combines short-term dependencies and long historical patterns through Transformer and TCN MLP module, respectively.
    \item \textbf{ASTGCN}~\cite{astgcn} incorporates weekly and daily data with attention mechanism to extract periodic patterns.
\end{itemize}


\newpage
\section*{NeurIPS Paper Checklist}

\begin{enumerate}

\item {\bf Claims}
    \item[] Question: Do the main claims made in the abstract and introduction accurately reflect the paper's contributions and scope?
    \item[] Answer: \answerYes{} 
    \item[] Justification: The claims are consistent with the contributions and scope as detailed in the abstract and introduction.

\item {\bf Limitations}
    \item[] Question: Does the paper discuss the limitations of the work performed by the authors?
    \item[] Answer: \answerYes{}{} 
    \item[] Justification: The existing limitation of our work is given in the appendix.

\item {\bf Theory assumptions and proofs}
    \item[] Question: For each theoretical result, does the paper provide the full set of assumptions and a complete (and correct) proof?
    \item[] Answer: \answerNA{} 
    \item[] Justification: The paper does not include theoretical claim.

    \item {\bf Experimental result reproducibility}
    \item[] Question: Does the paper fully disclose all the information needed to reproduce the main experimental results of the paper to the extent that it affects the main claims and/or conclusions of the paper (regardless of whether the code and data are provided or not)?
    \item[] Answer: \answerYes{} 
    \item[] Justification: The experimental setup and datasets are detailed in Section 4 and the proposed architecture is outlined in Section 3. Train/test details and model hyperparameters are included in the main text.

\item {\bf Open access to data and code}
    \item[] Question: Does the paper provide open access to the data and code, with sufficient instructions to faithfully reproduce the main experimental results, as described in supplemental material?
    \item[] Answer: \answerYes{} 
    \item[] Justification: Access to code is included in the introduction section.

\item {\bf Experimental setting/details}
    \item[] Question: Does the paper specify all the training and test details (e.g., data splits, hyperparameters, how they were chosen, type of optimizer) necessary to understand the results?
    \item[] Answer: \answerYes{} 
    \item[] Justification: Data description with train/val/test splits and hyperparameters are given in the main text.

\item {\bf Experiment statistical significance}
    \item[] Question: Does the paper report error bars suitably and correctly defined or other appropriate information about the statistical significance of the experiments?
    \item[] Answer: \answerYes{} 
    \item[] Justification: The standard deviation error on the results is given in the appendix.

\item {\bf Experiments compute resources}
    \item[] Question: For each experiment, does the paper provide sufficient information on the computer resources (type of compute workers, memory, time of execution) needed to reproduce the experiments?
    \item[] Answer: \answerYes{} 
    \item[] Justification: Time of inference and test is given in the appendix.
    
\item {\bf Code of ethics}
    \item[] Question: Does the research conducted in the paper conform, in every respect, with the NeurIPS Code of Ethics \url{https://neurips.cc/public/EthicsGuidelines}?
    \item[] Answer: \answerYes{} 
    \item[] Justification: The research adheres to the NeurIPS Code of Ethics.

\item {\bf Broader impacts}
    \item[] Question: Does the paper discuss both potential positive societal impacts and negative societal impacts of the work performed?
    \item[] Answer: \answerYes{}{} 
    \item[] Justification: Our work can positively contribute to smarter and more reliable transportation systems, by improving robustness under irregular and dynamically changing traffic patterns.
    
\item {\bf Safeguards}
    \item[] Question: Does the paper describe safeguards that have been put in place for responsible release of data or models that have a high risk for misuse (e.g., pre-trained language models, image generators, or scraped datasets)?
    \item[] Answer: \answerNA{} 
    \item[] Justification: The paper poses no such risk.

\item {\bf Licenses for existing assets}
    \item[] Question: Are the creators or original owners of assets (e.g., code, data, models), used in the paper, properly credited and are the license and terms of use explicitly mentioned and properly respected?
    \item[] Answer: \answerYes{}{} 
    \item[] Justification: The original creators have been attributed in the form of citations.

\item {\bf New assets}
    \item[] Question: Are new assets introduced in the paper well documented and is the documentation provided alongside the assets?
    \item[] Answer: \answerNA{} 
    \item[] Justification: No new assets are included in the paper.

\item {\bf Crowdsourcing and research with human subjects}
    \item[] Question: For crowdsourcing experiments and research with human subjects, does the paper include the full text of instructions given to participants and screenshots, if applicable, as well as details about compensation (if any)? 
    \item[] Answer: \answerNA{} 
    \item[] Justification: The research does not involve crowdsourcing experiments or research with human objects.

\item {\bf Institutional review board (IRB) approvals or equivalent for research with human subjects}
    \item[] Question: Does the paper describe potential risks incurred by study participants, whether such risks were disclosed to the subjects, and whether Institutional Review Board (IRB) approvals (or an equivalent approval/review based on the requirements of your country or institution) were obtained?
    \item[] Answer: \answerNA{}{} 
    \item[] Justification: The paper does not involve research with human subjects.

\item {\bf Declaration of LLM usage}
    \item[] Question: Does the paper describe the usage of LLMs if it is an important, original, or non-standard component of the core methods in this research? Note that if the LLM is used only for writing, editing, or formatting purposes and does \emph{not} impact the core methodology, scientific rigor, or originality of the research, declaration is not required.
    \item[] Answer: \answerNA{} 
    \item[] Justification: The core method development in this work does not involve LLMs as any important, original or non-standard components.

\end{enumerate}

\end{document}